\documentclass{article}

\usepackage[utf8]{inputenc}
\usepackage[T1]{fontenc}
\usepackage{microtype}
\usepackage{graphicx}
\usepackage[font=footnotesize]{caption}
\usepackage{subcaption}
\usepackage{booktabs} 
\usepackage{amsmath}
\usepackage{amssymb}
\usepackage{amsfonts}
\usepackage{nicefrac}
\usepackage[table]{xcolor}
\usepackage{multirow}
\usepackage{rotating}
\usepackage{siunitx}
\usepackage{xspace}
\usepackage{url}

\usepackage{hyperref}

\usepackage[preprint]{icml2026}

\usepackage{amsmath,amsfonts,bm}

\def\secref#1{section~\ref{#1}}

\def\eqref#1{equation~\ref{#1}}

\def\1{\bm{1}}

\DeclareMathAlphabet{\mathsfit}{\encodingdefault}{\sfdefault}{m}{sl}
\SetMathAlphabet{\mathsfit}{bold}{\encodingdefault}{\sfdefault}{bx}{n}

\def\model{TimeSqueeze\xspace}

\icmltitlerunning{\model: Dynamic Patching for Efficient Time Series Forecasting}

\begin{document}

\twocolumn[
  \icmltitle{\model: Dynamic Patching for Efficient Time Series Forecasting}

 \begin{icmlauthorlist}
  \icmlauthor{Sravan Kumar Ankireddy}{utex,comp}
  \icmlauthor{Nikita Seleznev}{comp}
  \icmlauthor{Nam H. Nguyen}{comp}
  \icmlauthor{Yulun Wu}{comp}
  \icmlauthor{Senthil Kumar}{comp}
  \icmlauthor{Furong Huang}{umd}
  \icmlauthor{C. Bayan Bruss}{comp}
\end{icmlauthorlist}

\icmlaffiliation{utex}{University of Texas at Austin, Austin, TX, USA}
\icmlaffiliation{comp}{Capital One, USA}
\icmlaffiliation{umd}{University of Maryland, College Park, MD, USA}

\icmlcorrespondingauthor{Sravan Kumar Ankireddy}{sravan.ankireddy@utexas.edu}


  \vskip 0.3in
]

\printAffiliationsAndNotice{Work done in part during an internship at Capital One.}

\begin{abstract}

Transformer-based time series foundation models face a fundamental trade-off in choice of tokenization: point-wise embeddings preserve temporal fidelity but scale poorly with sequence length, whereas fixed-length patching improves efficiency by imposing uniform boundaries that may disrupt natural transitions and blur informative local dynamics. In order to address these limitations, we introduce \model, a \emph{dynamic patching} mechanism that adaptively selects patch boundaries \emph{within each sequence} based on local signal complexity.
\model first applies a lightweight state-space encoder to extract full-resolution point-wise features, then performs content-aware segmentation by allocating \emph{short} patches to information-dense regions and \emph{long} patches to smooth or redundant segments. This variable-resolution compression preserves critical temporal structure while substantially reducing the token sequence presented to the Transformer backbone. Specifically for large-scale pretraining, \model attains up to $20\times$ faster convergence and $8\times$ higher data efficiency compared to equivalent point-token baselines.
Experiments across long-horizon forecasting benchmarks show that, \model consistently outperforms comparable architectures that use either point-wise tokenization or fixed-size patching. 

\end{abstract}

% ...





\section{Introduction}\label{sec:intro}

Accurate time-series forecasting is crucial across numerous domains, including energy, finance, climate, and healthcare. Historically, forecasting has relied on narrow, task-specific statistical models; however, recent advances in deep learning have enabled the development of versatile, generalist models capable of cross-domain transfer. In particular, time-series foundation models trained on heterogeneous datasets offer flexible zero-shot and few-shot generalization across a wide range of forecasting tasks. 

Effective pretraining of these foundation models necessitates modeling long historical contexts, often extending to thousands of timesteps, which creates formidable computational and memory constraints. Recent studies demonstrate that increasing context length during pretraining yields substantial improvements in downstream inference performance
~\citep{gao2024train,liu2024timer}. Therefore, designing architectures that remain scalable and computationally efficient under long-context regimes is imperative for realizing the full potential of time series foundation models.

Central to addressing these scalability challenges is the design of an efficient tokenizer that effectively represents input signals in an embedding space while managing computational complexity. Current approaches predominantly adopt one of two strategies. The first approach involves independently encoding each time point~\citep{zhou2021informer,wu2021autoformer,zhou2022fedformer,ansari2024chronos,shi2024time}, which preserves fine-grained temporal variations and accommodates data of arbitrary frequency and seasonality. However, this point-wise encoding strategy suffers from limited scalability as sequence length increases, which is precisely the bottleneck that impedes long-context pretraining. The second approach, pioneered by~\citet{nie2022time} and subsequently adopted by numerous transformer-based forecasting models ~\citep{goswami2024moment,das2024decoder,woo2024unified,liu2024timer, liu2025sundial}, employs fixed-size patching to compress multiple consecutive time points into a single embedding. While this patching strategy significantly enhances computational scalability, it introduces some limitations that compromise its effectiveness. First, determining the optimal patch size is non-trivial and heavily dependent on dataset-specific characteristics such as sampling frequency and seasonal patterns, typically requiring empirical evaluation across different patch sizes for each dataset. Second, and perhaps more critically, many time series exhibit heterogeneous information density across different temporal regions, with some segments displaying rapid variations while others remaining relatively stable. This temporal heterogeneity renders uniform patching suboptimal, as it fails to adapt the representational granularity to the local complexity of the signal.

Motivated by these requirements, we propose \model, a hybrid time-series tokenization that combines the expressive power of point-embeddings with the computational efficiency of patch-embeddings. First, a lightweight state space model (SSM)~\cite{gu2023mamba} based encoder extracts local fine-grained features at full resolution. Then, a dynamic patching module groups these embeddings into patches of varying sizes, allocating smaller patches to information-rich regions and larger patches to redundant ones, yielding a variable-resolution representation. This results in significantly fewer tokens for the transformer backbone while preserving salient temporal dynamics, thereby overcoming fixed patch size limitations and enabling scalable, high-fidelity modeling.



Our contributions are as follows:
\begin{itemize}

    \item We propose \model, a hybrid tokenizer with an efficient SSM-based encoder--decoder that preserves fine-grained long-context features while enabling content-aware, truly dynamic patching via adaptive downsampling.
  
    \item We establish that \model integrates seamlessly with various Transformer backbones, enabling pretraining of large-scale time series foundation models with substantially reduced training budgets.

    \item We demonstrate that \model achieves performance on par with state-of-the-art point embedding models while delivering up to $20\times$ faster training and $8\times$ higher pre-training data efficiency.

    \item We evaluate \model across diverse backbones and pretraining datasets, showing consistent gains over prior patching and tokenization methods in both zero-shot and full-shot settings on univariate and multivariate forecasting tasks.

  
\end{itemize}

\section{Related Works. }


\textbf{Long-sequence architectures.}  
While Transformer architectures~\citep{vaswani2017attention} have shown strong time series forecasting performance due to their expressivity and flexibility, their quadratic computational and memory complexity with respect to sequence length limits their scalability to long historical contexts. Innovations such as~\citep{li2019enhancing, wu2021autoformer, zhou2021informer} have adapted Transformers for long-term forecasting, but pretraining on extremely long contexts remains challenging. Recently, time-series foundation models have demonstrated scalability to long contexts supporting arbitrary forecasting horizons, while Time-MoE~\citep{shi2024time} leveraged Mixture-of-Experts routing to enable the first billion-parameter model with tractable inference. Despite these advances, the cost of long-context pretraining remains high due to the underlying Transformer backbone. Although SSM architectures handle long contexts more efficiently, they remain underexplored for time series forecasting, highlighting the need for scalable methods for efficient long-context processing. 

\textbf{Patch-based tokenization.}  
Introduced in PatchTST~\citep{nie2022time}, patch-based compression has emerged as a fundamental technique for scaling time series foundation models. By embedding contiguous sub-sequences (patches) rather than individual time points, this approach reduces the effective sequence length while preserving essential local temporal patterns. Subsequent foundation models, including TimesFM~\citep{das2024decoder}, Moment~\citep{goswami2024moment}, Moirai~\citep{woo2024unified}, and Timer-XL~\citep{liu2024timer}, have adopted this paradigm, collectively demonstrating that patching enables more efficient training and inference.
However, these approaches utilize a fixed patch size for a given sequence, limiting their application to real-world data with high temporal variance,
underscoring the need for dynamic, data-driven compression strategies that can adjust patching to varying temporal structures within a series. 

\textbf{Dynamic patching in time-series modeling.} 
Several recent methods explored the idea of adapting the patch size based on input automatically, instead of hand-crafted heuristics. 
HDMixer~\cite{huang2024hdmixer} enables extension of patch size by selectively combining adjoining fixed-size patches. IMTS~\cite{zhang2024irregular} is tailored to irregularly sampled series, where the patch size is adjusted dynamically in order to maintain the same number of temporal observations within each patch. LightGTS~\cite{wang2025lightgts} adjusts the patch size for each input signal based on periodicity in the Fourier domain and SRSNet~\cite{wu2025enhancing} introduces a dynamic selection and combining of features to improve the forecasting capability. 
However, none of these methods perform \textbf{within-sequence dynamic patching} that relocates patch boundaries and varies patch sizes inside each series based on local signal statistics. More recently, EntroPE~\cite{abeywickrama2025entrope} introduces entropy based patching of the input time series but requires discretizing the input and learning a separate model for entropy calculation, adding additional overhead.

\textbf{Insights from language modeling.}  
Similar challenges arise in large language models (LLMs), where the choice of input representation has a direct impact on scalability and fidelity. Conventional tokenization introduces systematic biases and brittle dependencies, motivating tokenizer-free models that operate at the byte level. Yet, naïve byte-level processing leads to prohibitively long input sequences \citep{slagle2024spacebyte}, straining attention-based architectures. To overcome this, adaptive compression techniques have been proposed. The Byte Latent Transformer (BLT) dynamically merges predictable byte spans into compact latent tokens using entropy-guided segmentation \citep{pagnoni2024byte}, while H-Net \citep{hwang2025dynamic}, inspired by U-Net \citep{ronneberger2015u} and its broad adaptation in vision \citep{childvery,ho2020denoising,wucounterfactual}, compresses and reconstructs sequences in various resolutions, and uses a state-space model for more efficient byte-level processing. These approaches highlight a key principle: efficiency and accuracy can be jointly achieved by allocating higher granularity to information-dense regions and applying more aggressive compression where redundancy dominates.

\section{Methodology}\label{sec:methodology}

 In this work, we will mainly focus on the task of designing an efficient tokenizer for time-series forecasting foundation models and refer to the combined model as \model from here on for simplicity.
 
\noindent\textbf{Problem Statement.} The fundamental objective in time-series forecasting is to predict future values based on historical observations. Given a sequence of $T$ historical data points, $\mathbf{X}_{1:T} = (x_1, x_2, \ldots, x_T) \in \mathbb{R}^T$, the goal is to estimate the next $H$ values of the series. This is formalized via a model $f_{\theta}$ that maps the historical context to future predictions, i.e., $\hat{\mathbf{X}}_{T+1:T+H} = f_{\theta}(\mathbf{X}_{1:T}) \in \mathbb{R}^H$. Adopting the channel independence principle of~\citet{nie2022time}, the model can flexibly process multivariate time series by decomposing inputs into collections of univariate series. This general formulation enables time-series foundation models to address forecasting tasks with arbitrary input dimensionality, thereby supporting broad applicability across diverse, real-world domains. Note that we also consider explicit modeling of\textbf{ multivariate} forecasting in~\secref{sec:gen}.

\begin{figure*}
    \centering
    \includegraphics[width=0.9\textwidth]{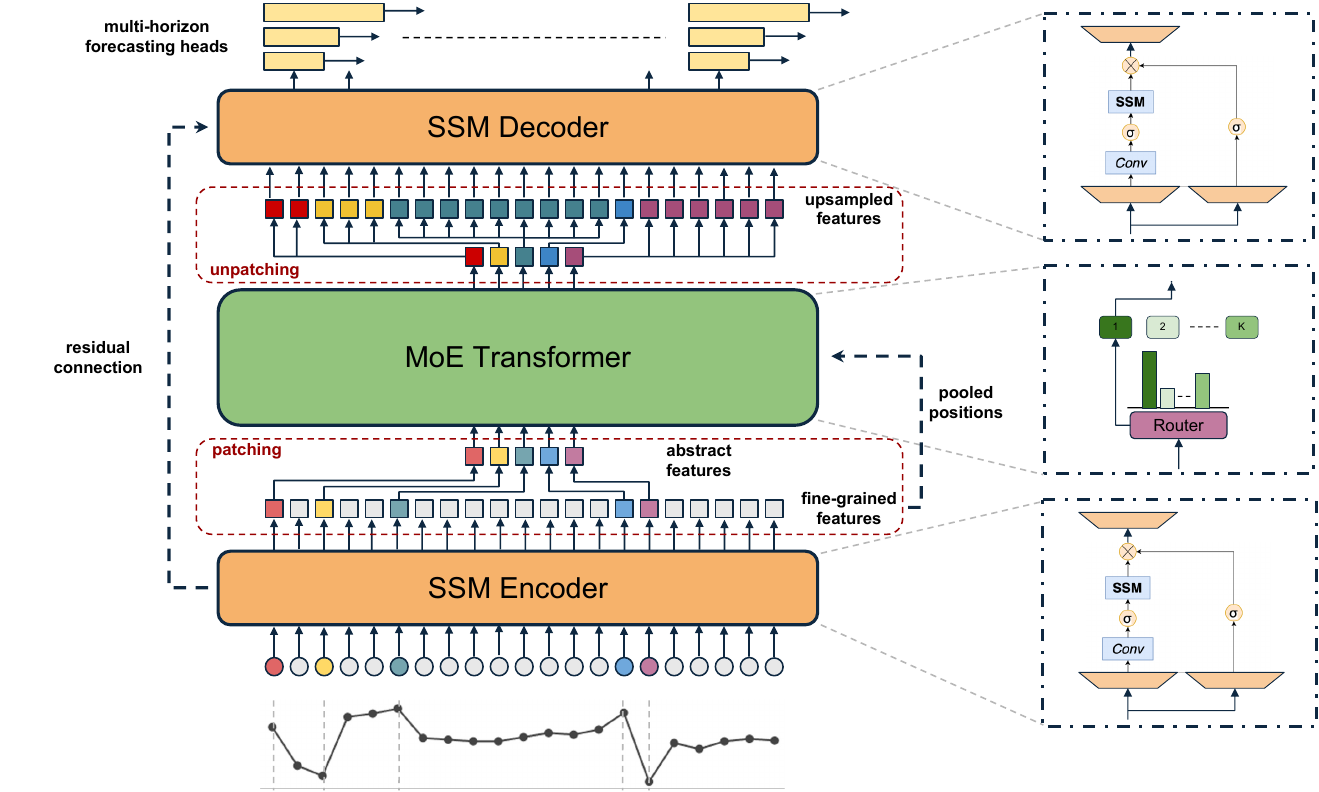}

          \caption{Architectural overview of \model forecasting model. An SSM encoder first processes the raw series at full resolution to extract fine-grained features. Dynamic patching then adaptively compresses the sequence, selecting the salient subset of features. A Transformer backbone performs contextual modeling on the downsampled features, and an unpatching module upsamples the signal to the original resolution while preserving causality. Finally, an SSM decoder combines the compressed and fine-grained features, passing the hybrid features to multi-horizon heads, thereby improving efficiency without sacrificing temporal fidelity.}

    \label{fig:arch}
    \vspace{-1.5em}
\end{figure*}

\subsection{Architectural Overview}\label{sec:arch}

To combine the expressivity of point-embeddings with the computational efficiency of patch embeddings, \model employs a hybrid multi-resolution architecture with four key components: (1) a lightweight encoder-decoder pair operating at full input resolution to capture fine-grained local features, (2) adaptive patching modules that dynamically compute salient features for efficient downsampling and upsampling, (3) a decoder-only mixture-of-experts (MoE) Transformer backbone for modeling causal dependencies at scale, and (4) a multi-horizon forecasting head that jointly optimizes predictions across multiple time horizons to support both short- and long-term forecasting, as shown in Figure~\ref{fig:arch}.
The first two components together form the hybrid tokenization mechanism in \model. 

Formally, the end-to-end model can be described as
\begin{equation}
\begin{aligned}
H_{1:T} &= \mathcal{E}(X_{1:T}), \quad
Z_{1:P} = \mathcal{M}\!\left(\mathcal{P}(H_{1:T})\right), \\
Y_{1:T} &= \mathcal{D}\!\left(H_{1:T}, \mathcal{U}(Z_{1:P})\right)
\end{aligned}
\end{equation}
where $\mathcal{E}$ is the encoder, $\mathcal{P}$ is the patching module, 
$\mathcal{M}$ is the MoE Transformer backbone, $\mathcal{U}$ is the unpatching module, 
and $\mathcal{D}$ is the decoder. 
Here, $X_{1:T} \in \mathbb{R}^T$ denotes the original input sequence, 
$H_{1:T} \in \mathbb{R}^{T \times D}$ denotes the $D$-dimensional encoder embeddings, 
$Z_{1:P} \in \mathbb{R}^{P \times D}$ denotes the patch-level latent representation after $\mathcal{P}$ and $\mathcal{M}$, 
and $Y_{1:T} \in \mathbb{R}^{T \times D}$ denotes the decoder embeddings that serve as the final representation 
for downstream forecasting.

\subsubsection{State-Space Encoder and Decoder}
The encoder and decoder modules operate directly on input time series at native resolution to preserve fine-grained temporal details essential for accurate forecasting, particularly in high-frequency data. To handle long, uncompressed sequences efficiently while generating representations suitable for subsequent patching, both modules are constructed using Mamba layers~\citep{gu2023mamba}. 

Mamba offers nearly linear computational scaling with respect to sequence length, enabling extraction of intricate local patterns from extended contexts by the encoder, without the quadratic complexity of traditional Transformer architectures. Further, the decoder uses the same architecture to efficiently combines outputs from the Transformer backbone with residual embeddings from the encoder to produce final representations for forecasting, creating a rich multi-scale feature space that captures both local fine-grained patterns and global contextual dependencies.




\subsubsection{Dynamic Patching and Unpatching}  
After the encoder produces fine-grained representations, the patching module compresses the sequence of embeddings before passing them to the Transformer backbone. The objective is to allocate computational resources efficiently by employing a dynamic patching strategy that adapts to the local complexity of the input signal. This strategy forms larger patches to compress regions of low information density while using smaller patches to preserve detail in regions of high information content
(Appendix~\ref{sec:vis_patches}).

\textbf{Patching.} Unlike language models that operate on discrete token sequences, time series data often exist in continuous space and exhibit rich statistical properties. This continuous nature makes time series particularly amenable to characterization via statistical measures such as local variance or power, without relying on external metrics for guidance~\citep{pagnoni2024byte}.
We leverage this by tracking the absolute difference between consecutive samples, comparing it to the average signal power within a predetermined lookback window, and then computing the patch boundaries in the original signal space rather than the embedding space. Formally, we maintain a sliding window \(\mathcal{W}_i = \{x_{i-L}, \ldots, x_{i-1}\}\) of length \(L\) to compute the local average power as 
\[
P_i = \frac{1}{L} \sum_{j=i-L}^{i-1} x_j^2.
\]
Our adaptive patching mechanism declares a patch boundary at timestep \(i\) if the absolute difference between consecutive samples exceeds a threshold scaled by the local power, which we refer to as \textbf{relative deviation-based patching}:
\[
b_i = \begin{cases} 
1 & \text{if } |x_i - x_{i-1}| > \tau \sqrt{P_i} \\
0 & \text{otherwise}
\end{cases}.
\]
Here, \(\tau > 0\) is a tunable threshold parameter controlling patch sensitivity and the average compression ratio. Using \(\sqrt{P_i}\) normalizes the threshold with respect to signal amplitude, allowing the method to adapt dynamically across varying signal magnitudes and variances. Intuitively, can be thought of as locating the \textbf{points of significant local change}, where a new patch should begin so that the SSM embedding/token at the patch boundary \textbf{can adequately summarize} the upcoming region.
Once patch boundaries are determined, the embeddings within each patch are compressed by retaining only the boundary embeddings and discarding intermediate ones (Figure~\ref{fig:arch}). Note that retaining only the boundary embeddings helps preserve causality for the subsequent unpatching step.

\textbf{Unpatching.} The unpatching module restores the compressed embeddings to the original sequence length while maintaining causal consistency. After backbone processing of boundary embeddings, each updated embedding is repeated across all timesteps within its corresponding patch and passed to the decoder for upsampling. Since boundary embeddings represent the start of each patch, the reconstructed output at timestep \(t\) depends only on inputs from times \(\leq t\), preventing leakage of future information. 

\textbf{Positional Information.} Unlike language models, which predict the next discrete token, time series foundation models demand more nuanced objective during pretraining. For instance, forecasting must occur at a specified frequency within the original continuous signal space, not within the compressed embedding space. Prior works on tokenizer-free language modeling, such as BLT~\citep{pagnoni2024byte} and Dynamic Chunking~\citep{hwang2025dynamic}, do not retain the original positional indices and restrict the attention mechanism to relative positional information post-downsampling. In contrast, \model explicitly preserves the position IDs of embeddings before downsampling and utilizes these absolute positions to compute attention after compression. 

\subsubsection{Mixture-of-Experts Transformer Backbone}
Due to its modular design, our hybrid feature extraction framework is compatible with any existing time-series forecasting backbone. In this work, we adopt the Time-MoE backbone~\citep{shi2024time}, a scalable decoder-only Transformer augmented with a sparse MoE routing mechanism. 
Time-MoE incorporates several enhancements to improve training stability and forecast accuracy: it employs RMSNorm for layer normalization and replaces absolute positional encoding with Rotary Positional Embeddings (RoPE), facilitating better handling of variable sequence lengths and improved extrapolation.  
Following established design patterns, the standard feed-forward network (FFN) is replaced by an MoE layer containing a pool of $N$ non-shared experts alongside one shared expert that consolidates common knowledge. For each input token, a routing mechanism selects the top $K$ non-shared experts to process the signal, enabling efficient scaling to billions of parameters while maintaining manageable inference costs.

\subsubsection{Multi-horizon forecasting}  
To enhance forecasting flexibility and robustness, we employ a multi-horizon forecasting head as introduced in~\citep{shi2024time}. This approach enables simultaneous prediction across multiple future horizons rather than restricting the model to a single forecast length. Specifically, it consists of multiple single-layer FFNs, each dedicated to a distinct forecasting horizon. The model is trained using a composite loss aggregating errors from all horizons, which improves generalization. During inference, a simple scheduling strategy selects the appropriate horizon-specific output, enabling the model to produce forecasts of arbitrary length flexibly.

\subsection{Model Training}\label{sec:train}

\textbf{Pretraining Dataset.} Efficient pretraining of a foundation model necessitates a large and diverse dataset. For this purpose, we employ the Time-300B dataset~\citep{shi2024time}, a high-quality, open-access dataset composed of time series from numerous public sources across various sectors, including weather, transportation, and finance, which is further expanded with synthetic data. It consists of a broad range of frequencies, ranging from seconds to yearly, and a massive scale of over 300 billion time points, making it well-suited for pretraining large-scale models.


\textbf
{Loss Formulation.} Following \cite{shi2024time}, our training objective is a composite loss function that combines a primary forecasting loss with an auxiliary term for load balancing, which enables a fair comparison against the point-embedding baseline Time-MoE. The primary auto-regressive loss, $\mathcal{L}_{\text{ar}}$, is the Huber Loss~\citep{huber1992robust}, chosen for its robustness against outliers:
\begin{equation}
\mathcal{L}_{ar}(x_t, \hat{x}_t) =
\begin{cases}
    \frac{1}{2}(x_t - \hat{x}_t)^2, & \text{if } |x_t - \hat{x}_t| \leq \delta, \\
    \delta \left( |x_t - \hat{x}_t| - \frac{1}{2} \delta \right), & \text{otherwise,}
\end{cases}
\end{equation}
where $\delta$ is a hyperparameter that balances the quadratic ($L_2$) and linear ($L_1$) penalties.

To ensure balanced expert utilization and prevent routing collapse, we incorporate an auxiliary loss, $\mathcal{L}_{\text{aux}}$, as proposed in~\cite{fedus2022switch}:
\begin{equation}
\mathcal{L}_{\text{aux}} = N \sum_{i=1}^{N} f_i r_i,
\end{equation}
where $f_i$ is the fraction of tokens dispatched to expert $i$, and $r_i$ is the average router probability assigned to the expert. The final training loss, $\mathcal{L}$, averages the auto-regressive loss across $K$ multi-resolution projections and combines it with the weighted auxiliary loss:
\begin{equation}
\mathcal{L} = \frac{1}{K} \sum_{j=1}^{K} \mathcal{L}_{\text{ar}} \left( \mathbf{X}_{t+1:t+p_j}, \hat{\mathbf{X}}_{t+1:t+p_j} \right) + \alpha \mathcal{L}_{\text{aux}},
\end{equation}
where $p_j$ is the forecast horizon for the $j$-th projection and $\alpha$ is a scaling coefficient.

\textbf{Model Configuration.} We consider two model sizes in this work, demonstrating the scalability of our approach. \model$_{\text{base}}$ has a total of 117M parameters with 54M active parameters, while \model$_{\text{large}}$ contains 469M total parameters with 216M active parameters. Both models are trained for 100,000 steps with a batch size of 256 and a maximum context length of 2048, corresponding to 500K time points per iteration and a total of 50B time steps during pretraining. Finally, for the patching and unpatching modules, we target an average compression rate of 4 in \model by setting the threshold factor \(\tau = 0.3\), and limiting the maximum patch size to $8$, resulting in an average compression ratio of $4 \times$ on the pretraining dataset, balancing computational savings and information preservation. During inference, we use the same $\tau$ across all datasets, demonstrating the \textbf{robustness of $\tau$}. Further configuration details are provided in Appendix~\ref{sec:pretraining_conifg}.

\section{Experimental Results}\label{sec:results}

We aim to demonstrate that \model improves both efficiency and forecasting performance over point-embedding models by dynamically compressing the input context under a fixed \textbf{computational budget}. We adopt Time-MoE as our evaluation backbone due to the public availability of its pretraining pipeline and dataset. In our implementation, we keep the forecasting backbone and all training settings unchanged, and modify only the tokenization stage: we replace Time-MoE’s SwiGLU-based point-wise tokenizer with our SSM-based dynamic patching module. This controlled setup \textbf{isolates the effect of dynamic context compression} and enables a fair comparison between the two models.


\textbf{Baselines.}  We use Time-MoE as our baseline, and 
pretrain \model following the training scheme of~\cite{shi2024time}, but using $8 \times$ lesser data and $\approx 20 \times$ less train time, as shown in Figure~\ref{fig:pretraining}. We forecast on four prediction horizons $\{96, 192, 336, 720\}$ but use the same context length of $512$ in all cases.  
While, we study the point-forecasting performance of \model, but it can easily be extended to provide probabilistic forecasts by substituting the model's linear projection head with a probabilistic head. We  assess model performance using the mean squared error (MSE) and mean absolute error (MAE), computed between the predicted values and the ground truth.
For completeness, we also compare against Moirai-large~\citep{woo2024unified}, TimesFM~\citep{das2024decoder}, Moment~\citep{ansari2024chronos}, and Chronos~\citep{goswami2024moment}, with results taken from~\cite{shi2024time}.


\begin{table*}[ht]
\centering
\scriptsize 
\renewcommand{\arraystretch}{1.0} 
\setlength{\tabcolsep}{2pt} 
\sisetup{table-format=1.3} 
\caption{Performance comparison of zero-shot forecasting. Best results are highlighted in \textcolor{red}{red} and second-best in \textcolor{blue}{blue}. 
}
\vspace{1em}
\label{tab:results_zero_shot}
\resizebox{0.9\textwidth}{!}{
    \begin{tabular}{@{}ll *{8}{S S}@{}}
    \toprule
    \multicolumn{1}{c}{\multirow{2}{*}{\textbf{Models}}} & \multicolumn{1}{c|}{\multirow{2}{*}{\textbf{Metrics}}} & \multicolumn{2}{c}{\textbf{\model$_{\text{base}}$}} & \multicolumn{2}{c}{\textbf{\model$_{\text{large}}$}} & \multicolumn{2}{c}{\textbf{Time-MoE$_{\text{base}}$}} & \multicolumn{2}{c}{\textbf{Time-MoE$_{\text{large}}$}} & \multicolumn{2}{c}{\textbf{Moirai$_{\text{base}}$}} & \multicolumn{2}{c}{\textbf{TimesFM}} & \multicolumn{2}{c}{\textbf{Moment}} & \multicolumn{2}{c}{\textbf{Chronos$_{\text{large}}$}} \\
    \cmidrule(lr){3-4} \cmidrule(lr){5-6} \cmidrule(lr){7-8} \cmidrule(lr){9-10} \cmidrule(lr){11-12} \cmidrule(lr){13-14} \cmidrule(lr){15-16} \cmidrule(lr){17-18}
    & & \textbf{MSE} & \textbf{MAE} & \textbf{MSE} & \textbf{MAE} & \textbf{MSE} & \textbf{MAE} & \textbf{MSE} & \textbf{MAE} & \textbf{MSE} & \textbf{MAE} & \textbf{MSE} & \textbf{MAE} & \textbf{MSE} & \textbf{MAE} & \textbf{MSE} & \textbf{MAE} \\
    \midrule
    \multirow{5}{*}{ETTh1} & 96 & 0.359 & 0.385 & 0.360 & \textcolor{red}{0.379} & \textcolor{blue}{0.357} & 0.381 & \textcolor{red}{0.350} & 0.382 & 0.376 & 0.392 & 0.414 & 0.404 & 0.688 & 0.557 & 0.441 & 0.390 \\
    & 192 & 0.400 & \textcolor{blue}{0.410} & 0.402 & 0.407 & \textcolor{blue}{0.388} & 0.412 & \textcolor{red}{0.384} & 0.404 & 0.417 & 0.413 & 0.465 & 0.434 & 0.688 & 0.560 & 0.502 & 0.424 \\
    & 336 & \textcolor{blue}{0.420} & 0.423 & 0.423 & \textcolor{red}{0.412} & \textcolor{red}{0.411} & 0.430 & \textcolor{red}{0.411} & 0.434 & 0.433 & \textcolor{red}{0.428} & 0.503 & 0.456 & 0.675 & 0.563 & 0.576 & 0.467 \\
    & 720 & \textcolor{blue}{0.428} & \textcolor{blue}{0.446} & 0.441 & 0.448 & \textcolor{red}{0.427} & 0.455 & 0.449 & 0.477 & 0.447 & \textcolor{red}{0.444} & 0.511 & 0.481 & 0.683 & 0.585 & 0.835 & 0.583 \\
    \rowcolor{gray!20}& \textbf{Avg.} & \textcolor{blue}{0.402} & 0.416 & 0.407 & \textcolor{blue}{0.414} & \textcolor{red}{0.394} & 0.419 & \textcolor{blue}{0.400} & 0.420 & 0.417 & \textcolor{red}{0.419} & 0.473 & 0.443 & 0.683 & 0.566 & 0.588 & 0.466 \\
    \hline \\[-0.5em]
    \multirow{5}{*}{ETTh2} & 96 & \textcolor{red}{0.282} & 0.346 & \textcolor{blue}{0.290} & 0.355 & 0.305 & 0.359 & 0.302 & 0.354 & 0.294 & \textcolor{red}{0.330} & 0.315 & 0.349 & 0.342 & 0.396 & 0.320 & 0.345 \\
    & 192 & \textcolor{red}{0.349} & 0.394 & 0.368 & 0.413 & \textcolor{blue}{0.351} & 0.386 & 0.364 & 0.385 & 0.365 & \textcolor{red}{0.375} & 0.388 & 0.395 & \textcolor{blue}{0.354} & 0.402 & 0.406 & 0.399 \\
    & 336 & 0.379 & 0.422 & 0.405 & 0.447 & 0.391 & 0.418 & 0.417 & 0.425 & \textcolor{blue}{0.376} & \textcolor{red}{0.390} & 0.422 & 0.427 & \textcolor{red}{0.356} & 0.407 & 0.492 & 0.453 \\
    & 720 & 0.444 & 0.471 & 0.445 & 0.441 & 0.419 & 0.454 & 0.537 & 0.496 & \textcolor{blue}{0.416} & \textcolor{red}{0.433} & 0.443 & 0.454 & \textcolor{red}{0.395} & 0.434 & 0.603 & 0.511 \\
    \rowcolor{gray!20}& \textbf{Avg.} & 0.363 & 0.408 & 0.377 & 0.414 & 0.366 & 0.404 & 0.405 & 0.415 & \textcolor{blue}{0.362} & \textcolor{red}{0.382} & 0.392 & 0.406 & \textcolor{red}{0.361} & 0.409 & 0.455 & 0.427 \\
    \hline \\[-0.5em]
    \multirow{5}{*}{ETTm1} & 96 & 0.312 & 0.344 & \textcolor{red}{0.304} & \textcolor{red}{0.334} & 0.338 & 0.368 & \textcolor{blue}{0.309} & 0.357 & 0.363 & \textcolor{blue}{0.356} & 0.361 & 0.370 & 0.654 & 0.527 & 0.457 & 0.403 \\
    & 192 & 0.372 & 0.385 & \textcolor{blue}{0.358} & \textcolor{red}{0.367} & \textcolor{blue}{0.353} & 0.388 & \textcolor{red}{0.346} & 0.381 & 0.388 & \textcolor{blue}{0.375} & 0.414 & 0.405 & 0.662 & 0.532 & 0.530 & 0.450 \\
    & 336 & 0.435 & 0.425 & \textcolor{blue}{0.403} & \textcolor{red}{0.396} & \textcolor{blue}{0.381} & 0.413 & \textcolor{red}{0.373} & 0.408 & 0.416 & \textcolor{red}{0.392} & 0.445 & 0.429 & 0.672 & 0.537 & 0.577 & 0.481 \\
    & 720 & 0.547 & 0.494 & \textcolor{blue}{0.486} & \textcolor{blue}{0.444} & 0.504 & 0.493 & \textcolor{blue}{0.475} & 0.477 & 0.460 & \textcolor{red}{0.418} & 0.512 & 0.471 & 0.692 & 0.551 & 0.660 & 0.526 \\
    \rowcolor{gray!20}& \textbf{Avg.} & 0.417 & 0.412 & \textcolor{blue}{0.388} & \textcolor{red}{0.385} & 0.394 & 0.415 & \textcolor{red}{0.376} & 0.406 & 0.406 & \textcolor{red}{0.385} & 0.433 & 0.418 & 0.670 & 0.536 & 0.555 & 0.465 \\
    \hline \\[-0.5em]
    \multirow{5}{*}{ETTm2} & 96 & \textcolor{blue}{0.181} & 0.275 & \textcolor{red}{0.179} & 0.272 & 0.201 & 0.291 & 0.197 & 0.286 & 0.205 & \textcolor{blue}{0.273} & 0.202 & \textcolor{red}{0.270} & 0.260 & 0.335 & 0.197 & \textcolor{blue}{0.271} \\
    & 192 & \textcolor{red}{0.248} & 0.323 & 0.251 & 0.325 & 0.258 & 0.334 & \textcolor{blue}{0.250} & 0.322 & 0.275 & 0.316 & 0.289 & 0.321 & 0.289 & 0.350 & 0.254 & \textcolor{red}{0.314} \\
    & 336 & \textcolor{red}{0.310} & 0.363 & \textcolor{blue}{0.319} & 0.368 & 0.324 & 0.373 & 0.337 & 0.375 & 0.329 & \textcolor{blue}{0.350} & 0.360 & 0.366 & 0.324 & 0.369 & \textcolor{blue}{0.313} & \textcolor{red}{0.353} \\
    & 720 & 0.431 & 0.437 & 0.425 & 0.428 & 0.488 & 0.464 & 0.480 & 0.461 & 0.437 & 0.411 & 0.462 & 0.430 & \textcolor{red}{0.394} & \textcolor{red}{0.409} & \textcolor{blue}{0.416} & 0.415 \\
    \rowcolor{gray!20}& \textbf{Avg.} & \textcolor{red}{0.292} & 0.349 & \textcolor{blue}{0.294} & \textcolor{blue}{0.348} & 0.317 & 0.365 & 0.316 & 0.361 & 0.311 & \textcolor{red}{0.337} & 0.328 & 0.346 & 0.316 & 0.365 & 0.295 & \textcolor{red}{0.338} \\
    \hline \\[-0.5em]  
    \multirow{5}{*}{Weather} & 96 & 0.167 & 0.217 & 0.170 & 0.221 & \textcolor{blue}{0.160} & 0.214 & \textcolor{red}{0.159} & \textcolor{blue}{0.213} & 0.220 & 0.217 & {-} & {-} & 0.243 & 0.255 & 0.194 & 0.235 \\
    & 192 & 0.218 & 0.269 & 0.224 & 0.275 & \textcolor{red}{0.210} & \textcolor{blue}{0.260} & \textcolor{blue}{0.215} & 0.266 & 0.271 & 0.259 & {-} & {-} & 0.278 & 0.329 & 0.249 & 0.285 \\
    & 336 & \textcolor{blue}{0.278} & 0.315 & 0.292 & 0.326 & \textcolor{red}{0.274} & \textcolor{blue}{0.309} & 0.291 & 0.322 & 0.286 & 0.297 & {-} & {-} & 0.306 & 0.346 & 0.302 & 0.327 \\
    & 720 & \textcolor{blue}{0.364} & 0.372 & 0.409 & 0.396 & 0.418 & 0.405 & 0.419 & 0.400 & 0.373 & 0.354 & {-} & {-} & \textcolor{red}{0.350} & 0.374 & 0.372 & 0.378 \\
    \rowcolor{gray!20}& \textbf{Avg.} & \textcolor{red}{0.257} & 0.293 & 0.274 & 0.305 & \textcolor{blue}{0.265} & 0.297 & 0.271 & 0.300 & 0.287 & \textcolor{blue}{0.281} & {-} & {-} & 0.294 & 0.326 & 0.279 & 0.306 \\
    \midrule
    \rowcolor{gray!20}\textbf{Average} & & \textcolor{red}{0.346} & 0.376 & \textcolor{blue}{0.348} & \textcolor{blue}{0.373} & \textcolor{blue}{0.347} & 0.380 & 0.352 & 0.380 & 0.357 & \textcolor{red}{0.361} & 0.407 & 0.403 & 0.465 & 0.440 & 0.434 & 0.400 \\
    \midrule
    \bottomrule
    \end{tabular}
}
\vspace{-1em}
\end{table*}

\subsection{Zero-shot forecasting}

We compare the zero-shot performance of \model$_{\text{base}}$ and \model$_{\text{large}}$ against Time-MoE$_{\text{base}}$ and $_{\text{large}}$ on the well-studied long-term forecasting benchmarks~\citep{zhou2021informer} and the Weather data~\citep{wu2021autoformer}.
These datasets were not included in the Time-300B dataset and not used for training the \model. 
Detailed zero-shot forecasting results are presented in Table~\ref{tab:results_zero_shot}, demonstrating that \model performs remarkably well, achieving a performance \textbf{similar} to that of Time-MoE. Further results for higher compression rates are provided in Appendix~\ref{sec:rate_vs_mse}.

Additional comparisons for \model against Time-MoE are presented in Section~\ref{sec:fixed_context}. We note that the performance of \model$_{\text{large}}$ is slightly worse than \model$_{\text{base}}$ in some scenarios, likely due to the limited training budget.

\subsection{In-distribution forecasting}\label{sec:full_shot_avg}
We now measure the full-shot performance by finetuning \model on the train split of the same benchmarks. For finetuning, we choose a learning rate of 1e-4 and fine-tune the pretrained model for just one epoch. We compare the full-shot performance against~\cite{liu2023itransformer, wang2024timemixer, wu2022timesnet, nie2022time, zeng2023transformers, lin2024cyclenet, lin2025temporal}, in addition to the finetuned version of Time-MoE$_{\text{base}}$. As seen from Table~\ref{tab:results_full_shot_time_moe_avg}, \model still performs close to Time-MoE, and outperforms all other baselines considered. Complete results are provided in Appendix~\secref{sec:full_shot_full}.

\begin{table*}
\centering
\scriptsize 
\renewcommand{\arraystretch}{1.0} 
\setlength{\tabcolsep}{2pt} 
\sisetup{table-format=1.3} 
\caption{Performance comparison of full-shot forecasting. Best results are highlighted in \textcolor{red}{red} and second-best in \textcolor{blue}{blue}. 
}
\vspace{1em}
\label{tab:results_full_shot_time_moe_avg}
\resizebox{0.9\textwidth}{!}{
\begin{tabular}{@{}l *{9}{S S}@{}}
\toprule
\multicolumn{1}{c}{\multirow{2}{*}{\textbf{Dataset}}} & \multicolumn{2}{c}{\textbf{\model$_{\text{base}}$}} & \multicolumn{2}{c}{\textbf{Time-MoE$_{\text{base}}$}} & \multicolumn{2}{c}{\textbf{iTransformer}} & \multicolumn{2}{c}{\textbf{TimeMixer}} & \multicolumn{2}{c}{\textbf{TimesNet}} & \multicolumn{2}{c}{\textbf{PatchTST}} & \multicolumn{2}{c}{\textbf{DLinear}} & \multicolumn{2}{c}{\textbf{CycleNet$_{\text{MLP}}$}} & \multicolumn{2}{c}{\textbf{TQNet}} \\
\cmidrule(lr){2-3} \cmidrule(lr){4-5} \cmidrule(lr){6-7} \cmidrule(lr){8-9} \cmidrule(lr){10-11} \cmidrule(lr){12-13} \cmidrule(lr){14-15} \cmidrule(lr){16-17} \cmidrule(lr){18-19}
& \textbf{MSE} & \textbf{MAE} & \textbf{MSE} & \textbf{MAE} & \textbf{MSE} & \textbf{MAE} & \textbf{MSE} & \textbf{MAE} & \textbf{MSE} & \textbf{MAE} & \textbf{MSE} & \textbf{MAE} & \textbf{MSE} & \textbf{MAE} & \textbf{MSE} & \textbf{MAE} & \textbf{MSE} & \textbf{MAE} \\
\midrule
ETTh1 & \textcolor{blue}{0.398} & \textcolor{blue}{0.419} & \textcolor{red}{0.379} & \textcolor{red}{0.406} & 0.454 & 0.447 & 0.448 & 0.442 & 0.454 & 0.450 & 0.468 & 0.454 & 0.529 & 0.522 & 0.457 & 0.441 & 0.441 & 0.434 \\
ETTh2 & \textcolor{blue}{0.350} & \textcolor{blue}{0.393} & \textcolor{red}{0.346} & \textcolor{red}{0.386} & 0.383 & 0.406 & 0.364 & 0.395 & 0.414 & 0.496 & 0.386 & 0.406 & 0.942 & 0.683 & 0.388 & 0.409 & 0.378 & 0.402 \\
ETTm1 & 0.383 & \textcolor{blue}{0.386} & \textcolor{red}{0.345} & \textcolor{red}{0.381} & 0.407 & 0.409 & 0.381 & 0.395 & 0.400 & 0.405 & 0.387 & 0.400 & 0.513 & 0.495 & 0.379 & 0.396 & \textcolor{blue}{0.377} & 0.393 \\
ETTm2 & \textcolor{red}{0.259} & \textcolor{blue}{0.321} & 0.271 & 0.335 & 0.288 & 0.332 & 0.275 & 0.323 & 0.291 & 0.332 & 0.280 & 0.326 & 0.757 & 0.610 & \textcolor{blue}{0.266} & \textcolor{red}{0.314} & 0.277 & 0.323 \\
Weather & 0.243 & 0.279 & \textcolor{red}{0.236} & 0.275 & 0.250 & 0.280 & \textcolor{blue}{0.240} & \textcolor{blue}{0.271} & 0.259 & 0.286 & 0.258 & 0.280 & 0.258 & 0.315 & 0.243 & \textcolor{blue}{0.271} & 0.242 & \textcolor{red}{0.269} \\
\midrule
\rowcolor{gray!20}\textbf{Average} & \textcolor{blue}{0.327} & \textcolor{blue}{0.360} & \textcolor{red}{0.315} & \textcolor{red}{0.357} & 0.356 & 0.375 & 0.342 & 0.365 & 0.364 & 0.394 & 0.356 & 0.373 & 0.600 & 0.525 & 0.347 & 0.366 & 0.343 & 0.364 \\
\bottomrule
\end{tabular}
}
\vspace{-1em}
\end{table*}

\subsection{Efficiency Comparison}
We now compare the training and inference efficiency of \model$_{\text{base}}$ with the point-embedding baseline Time-MoE$_{\text{base}}$ model in terms of GPU hours and memory utilization. All experiments were conducted on $2\times$ NVIDIA A100 80GB GPUs.

\begin{figure}[hbt!]
\centering
\begin{subfigure}[b]{0.4\textwidth}
    \centering
    \includegraphics[width=\textwidth]{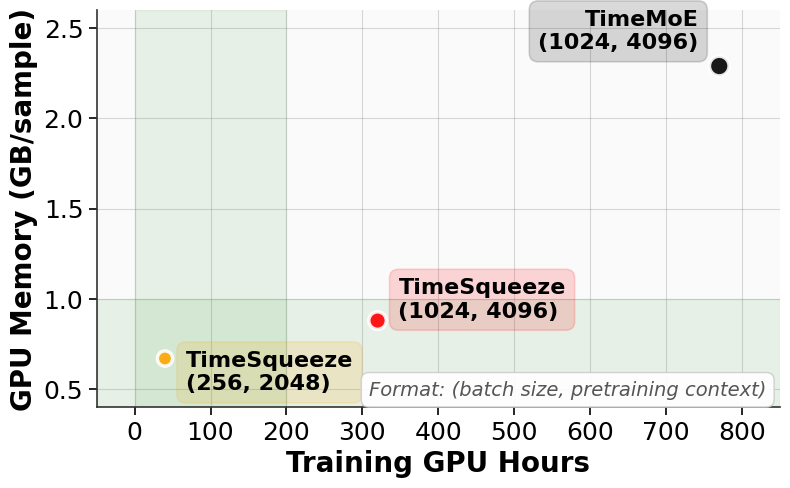}
    \subcaption{}
    \label{fig:pretraining}
\end{subfigure}
\hfill
\begin{subfigure}[b]{0.4\textwidth}
    \centering
    \includegraphics[width=\textwidth]{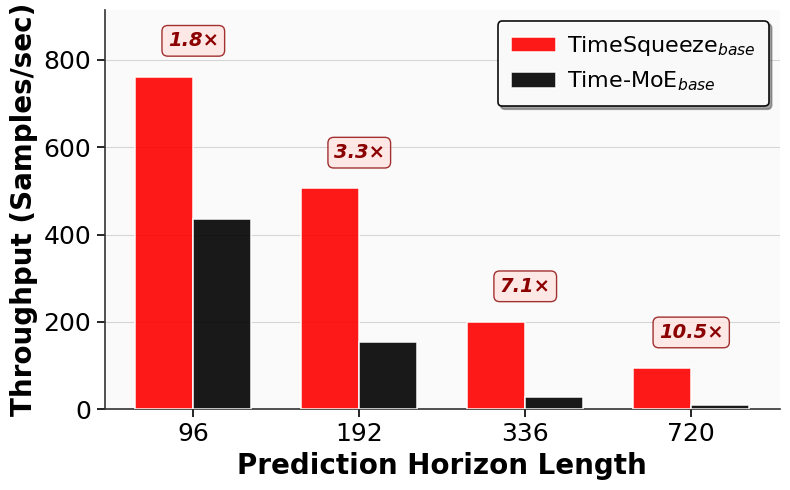}
    \subcaption{}
    \label{fig:throughput}
\end{subfigure}
\caption{Computational efficiency comparison between \model$_{\text{base}}$ and Time-MoE: (a) Training memory and time requirements across different batch sizes and context lengths. \model{} achieves comparable performance while reducing memory usage by $3.4\times$ and training time by $\approx 20\times$. (b) Inference throughput across prediction horizons. \model{} delivers up to $10.5\times$ higher throughput for longer prediction horizons.}
\label{fig:efficiency}
\vspace{-1em}
\end{figure}

In Figure~\ref{fig:pretraining}, we plot the pretraining time and memory required for different (batch size, context length) for Time-MoE and \model, when trained for $100,000$ iterations. When using $(1024, 4096)$, we see that \model uses $2.6 \times$ less memory and $2.4 \times$ less compute compared to Time-MoE. Furthermore, when running on a smaller budget, \model is trained with $(256, 2048)$, which uses $3.4 \times$ less memory and $19.25 \times$ less training time while still achieving performance comparable to Time-MoE, as shown in Table~\ref{tab:results_zero_shot}.

In Figure~\ref{fig:throughput}, we plot the inference throughput for different forecasting horizons. We use a context length of $512$ for \model and the original context lengths from~\citep{shi2024time} for Time-MoE. We see that \model scales more gracefully with respect to context length, showing up to $10.5 \times$ faster inference for longer prediction horizons, making \model more suitable for on-device inference.

\section{Ablation Studies}\label{sec:ablation}
We conduct systematic ablation studies to quantify the contributions of key components in \model. 
We use \model$_{\text{base}}$ for all ablation studies.
During inference, we use a context length of $512$ for \model and the original context lengths used in~\citep{shi2024time} for Time-MoE. 

\subsection{Model Components}\label{sec:componenets}

\emph{Dynamic vs. Fixed Patching.} We compare our proposed relative deviation-based dynamic patching approach with fixed patching. For the fixed patching baseline with patch size 4, embeddings are uniformly downsampled by retaining every 4th element. Results show that dynamic patching consistently outperforms fixed patching by effectively focusing computational resources on information-rich segments rather than optimizing only for a compression rate at the risk of discarding critical intermediate samples. This underscores the importance of dynamic compression strategies for handling temporal heterogeneity in time-series data.

\emph{Mamba vs. Linear Encoder.} To assess the importance of our SSM encoder-decoder, we replace it with simple linear embedding layers akin to the architectures used in Moirai~\citep{woo2024unified} and TimesFM~\citep{das2024decoder}. The SSM-based encoder achieves substantial gains over linear projections, confirming its suitability for capturing fine-grained temporal features and its inductive bias, which is beneficial for sequential compression.

\emph{Importance of Fine-Grained Features.} We evaluate the contribution of preserving detailed temporal information by ablating the residual connection illustrated in Figure~\ref{fig:arch}, relying solely on compressed features for forecasting. This modification results in noticeable performance degradation.

\emph{Positional Encoding Analysis.} We investigate the role of preserving positional information by comparing absolute position embeddings of boundary elements with relative positional encodings applied to compressed embeddings. Removing absolute positional cues results in notable performance drops, highlighting the necessity of absolute temporal positioning to maintain temporal coherence in the reconstructed sequences. 


\textbf{Observation.} Figure~\ref{fig:main_ablation} shows the summary of these ablations, by plotting the average MSE across the five benchmarking datasets for a prediction horizon of 96. 
The results clearly indicate that the inductive bias of SSM, combined with the dynamic context-aware pruning of SSM embeddings, is crucial to achieving optimal performance, while the residual connection and the use of absolute position IDs play a minor role. The full results are included in Appendix~\ref{app:add_ablation}, Table~\ref{tab:main_ablation}. Further experiments to study various training dynamics in detail are presented in detail in Appendix~\ref{sec:down},~\ref{sec:scale},~\ref{sec:rate_vs_mse},~\ref{sec:fixed_context}.

\subsection{Long-Context Pretraining}

Recent studies show that pretraining with longer context lengths can improve inference performance even when using shorter contexts during deployment~\citep{liu2024timer}. We investigate this by training \model with different maximum pretraining context lengths under a fixed token budget of $\approx$ 50B tokens. All models are trained for 100,000 steps, 
while maintaining an inference context length of 512 tokens.\looseness=-1

\begin{figure}[hbt!]
\centering
\begin{subfigure}[b]{0.4\textwidth}
    \centering
    \includegraphics[width=\textwidth]{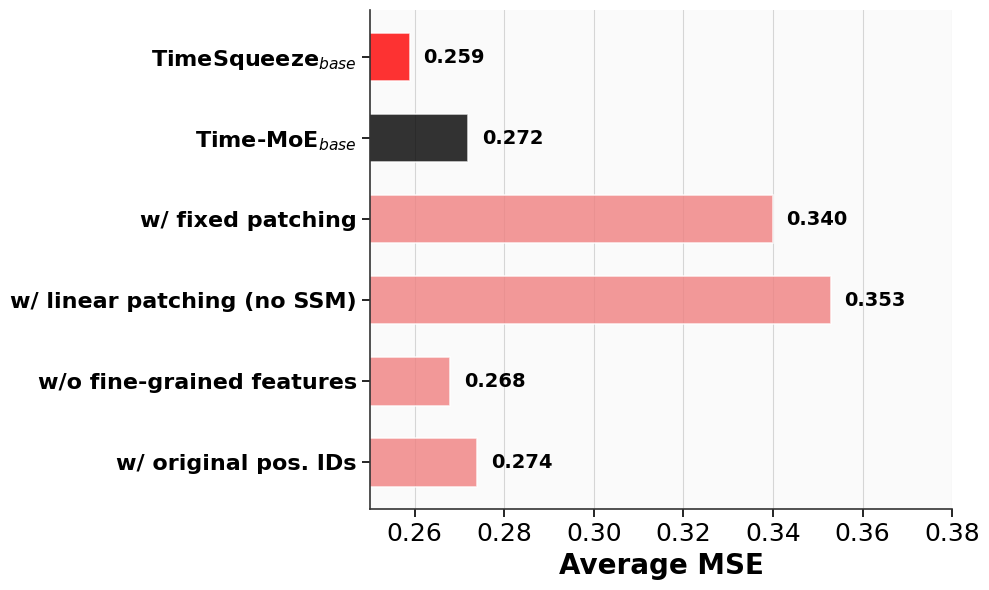}
    \subcaption{}
    \label{fig:main_ablation}
\end{subfigure}
\hfill
\begin{subfigure}[b]{0.4\textwidth}
    \centering
    \includegraphics[width=\textwidth]{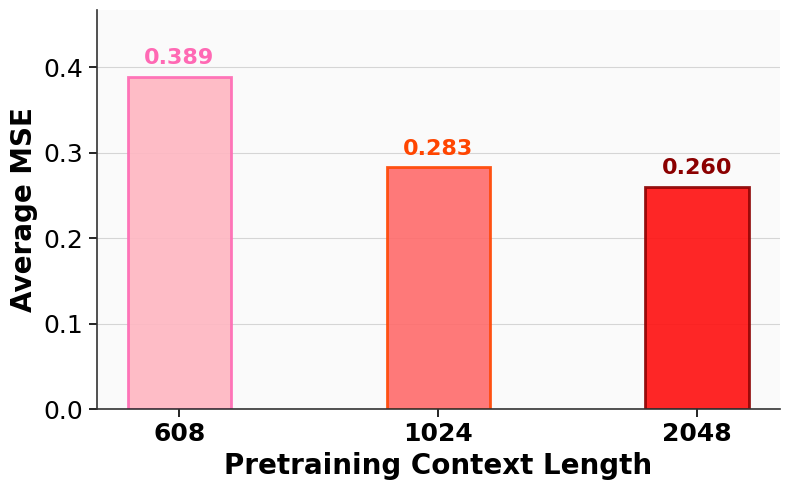}
    \subcaption{}
    \label{fig:pretrain_length_vs_performance}
\end{subfigure}
\caption{Ablation: (a) Average MSE across five benchmark datasets for prediction horizon 96 with different model components. (b) Pretraining Context Length vs Forecasting Performance: Longer pretraining context translates to improved performance, even when the inference context remains fixed at 512.}
\label{fig:combined_analysis}
\vspace{-1.5em}
\end{figure}

Figure~\ref{fig:pretrain_length_vs_performance} demonstrates that longer pretraining contexts consistently improve inference performance even when using a shorter inference context of 512 always. This indicates that exposure to extended sequences during pretraining enables \model to develop more robust temporal representations that effectively transfer to shorter inference contexts. Notably, unlike Time-MoE, \model achieves strong inference performance with short contexts, despite being pretrained on longer sequences, significantly reducing computational overhead during deployment (Appendix~\ref{sec:inference_short}).


\section{Generalization of \model}\label{sec:gen}


In~\secref{sec:results}, we considered decoder-only causal MoE transformer model. In order to show generalization of our tokenizer, we now consider a generic \textbf{encoder-only non-causal} transformer backbone without any MoE. Specifially, we adopt the \textit{GlobalTransformer} backbone from EntroPE~\cite{abeywickrama2025entrope} and perform \textbf{multi-variate forecasting} and compare against several lightweight forecasting models. 

For a fair comparison, we 
just replace the entropy based patching mechanism with \model based patching. We follow the exact same training and evaluation methodology. Table~\ref{tab:main_results} in Appendix~\ref{sec:full-shot-entrope} clearly shows that despite the change in backbone and training, \model still achieves competing performance and \textbf{outperforms} several state of the art in-distribution forecasting models, including \textbf{entropy-based patching} (EntroPE). We note that while hyperparameters have been tuned explicitly for EntroPE, we did not change them for \model and directly reuse them thus making the results presented potentially suboptimal.
Additional experiments to show generalization of \model gains across \textbf{forecasting backbone} and \textbf{pretraining datasets} can be found in Appendix~\secref{sec:gen_backbone} and~\secref{sec:gen_dataset}.

\section{Conclusion}\label{sec:conclusion}
We present \model, the first \textbf{truly dynamic patching} based tokenization mechanism to explore dynamic input compression, which combines the temporal fidelity of point embedding models with the computational efficiency of patch-based approaches \textbf{without relying on any external metrics}. 
Our relative deviation-based metric enables data-driven patching, producing representations that optimally allocate computational resources to where they provide the most significant benefit for forecasting.
\model achieves forecasting performance comparable to the baseline point embedding model Time-MoE, while achieving $8\times$ improvement in pretraining data efficiency and up to $20\times$ reduction in pretraining time. Across diverse backbones and zero-/full-shot univariate and multivariate forecasting benchmarks, \model \textbf{consistently outperforms equivalent models using alternative patching mechanisms}.




 Our work opens several promising research directions focusing on variable-rate patching and compression for time series forecasting. For instance, while relative threshold-based patching being \textbf{scale independent and robust}, it still requires hyperparameter tuning to achieve a target compression rate. Alternatively, patch boundaries could be learned end-to-end in embedding spaces~\citep{hwang2025dynamic} and natively to support variable compression rates.

\section*{Impact Statement}
This paper presents work whose goal is to advance the field of machine learning. There are many potential societal consequences of our work, none of which we feel must be specifically highlighted here.




\bibliography{references}
\bibliographystyle{icml2026}

\onecolumn

\newpage

\appendix

\section{In-distribution forecasting using encoder-only model}\label{sec:full-shot-entrope}

Results for experiments in~\secref{sec:gen} are presented below in Table~\ref{tab:main_results}.

\begin{table*}[ht]
\centering
\scriptsize
\caption{Multivariate time series forecasting results averaged across prediction horizons $T = \{96, 192, 336, 720\}$ with fixed input length $L = 96$ for all datasets. Best results are highlighted in \textcolor{red}{red} and second-best in \textcolor{blue}{blue}.}
\label{tab:main_results}
\resizebox{\textwidth}{!}{
\begin{tabular}{l|cc|cc|cc|cc|cc|cc}
\toprule
\multirow{2}{*}{Models} & \multicolumn{2}{c|}{ETTh1} & \multicolumn{2}{c|}{ETTh2} & \multicolumn{2}{c|}{ETTm1} & \multicolumn{2}{c|}{ETTm2} & \multicolumn{2}{c|}{Weather} & \multicolumn{2}{c}{Electricity} \\
& MSE & MAE & MSE & MAE & MSE & MAE & MSE & MAE & MSE & MAE & MSE & MAE \\
\midrule
Autoformer [2021] & 0.496 & 0.487 & 0.450 & 0.459 & 0.588 & 0.517 & 0.327 & 0.371 & 0.338 & 0.382 & 0.227 & 0.338 \\
FEDformer [2022] & 0.498 & 0.484 & 0.437 & 0.449 & 0.448 & 0.452 & 0.305 & 0.349 & 0.309 & 0.360 & 0.214 & 0.327 \\
DLinear [2023] & 0.461 & 0.457 & 0.563 & 0.519 & 0.404 & 0.408 & 0.354 & 0.402 & 0.265 & 0.315 & 0.225 & 0.319 \\
TimesNet [2023] & 0.495 & 0.450 & 0.414 & 0.427 & 0.400 & 0.406 & 0.291 & 0.333 & 0.251 & 0.294 & 0.193 & 0.304 \\
PatchTST [2023] & 0.516 & 0.484 & 0.391 & 0.411 & 0.406 & 0.407 & 0.290 & 0.334 & 0.265 & 0.285 & 0.216 & 0.318 \\
Time-FFM [2024] & 0.442 & 0.434 & 0.382 & 0.406 & 0.399 & 0.402 & 0.286 & 0.332 & 0.270 & 0.288 & 0.216 & 0.299 \\
HDMixer [2024] & 0.448 & 0.437 & 0.384 & 0.407 & 0.396 & 0.402 & 0.286 & 0.331 & 0.253 & 0.285 & 0.205 & 0.295 \\
iTransformer [2024] & 0.454 & 0.447 & 0.383 & 0.407 & 0.407 & 0.410 & 0.288 & 0.332 & 0.258 & 0.278 & \textcolor{blue}{0.178} & \textcolor{blue}{0.270} \\
TimeMixer [2024] & 0.459 & 0.444 & 0.390 & 0.409 & 0.382 & 0.397 & \textcolor{red}{0.279} & 0.324 & 0.245 & 0.276 & 0.182 & 0.272 \\
TimeBase [2025] & 0.463 & 0.429 & 0.409 & 0.425 & 0.431 & 0.420 & 0.290 & 0.332 & 0.252 & 0.279 & 0.227 & 0.296 \\
LangTime [2025] & 0.437 & \textcolor{red}{0.425} & 0.375 & 0.392 & 0.397 & \textcolor{blue}{0.392} & 0.284 & \textcolor{red}{0.321} & 0.252 & \textcolor{red}{0.273} & 0.201 & 0.285 \\
TimeKAN [2025] & \textcolor{blue}{0.418} & \textcolor{blue}{0.427} & 0.391 & 0.410 & 0.380 & 0.398 & 0.285 & 0.331 & 0.244 & \textcolor{red}{0.273} & 0.197 & 0.286 \\
FilterTS [2025] & 0.440 & 0.432 & 0.375 & 0.399 & 0.386 & 0.397 & \textcolor{red}{0.279} & \textcolor{blue}{0.323} & 0.253 & 0.280 & 0.184 & 0.275 \\
CALF [2025] & 0.441 & 0.435 & 0.372 & 0.395 & 0.396 & \textcolor{red}{0.391} & \textcolor{blue}{0.280} & \textcolor{red}{0.321} & 0.250 & \textcolor{blue}{0.274} & \textcolor{red}{0.177} & \textcolor{red}{0.266} \\
EntroPE [2025] & \textcolor{red}{0.416} & \textcolor{red}{0.425} & \textcolor{blue}{0.366} & \textcolor{blue}{0.387} & \textcolor{red}{0.378} & \textcolor{red}{0.391} & 0.286 & 0.335 & \textcolor{red}{0.242} & \textcolor{red}{0.273} & 0.182 & 0.271 \\
\textbf{TimeSqueeze (Ours)} & \textbf{0.432} & \textbf{0.436} & \textcolor{red}{\textbf{0.347}} & \textcolor{red}{\textbf{0.386}} & \textcolor{blue}{\textbf{0.379}} & \textbf{0.395} & \textbf{0.284} & \textbf{0.332} & \textcolor{blue}{\textbf{0.243}} & \textcolor{blue}{\textbf{0.274}} & \textbf{0.196} & \textbf{0.288} \\
\bottomrule
\end{tabular}
}
\end{table*}


\section{Pretraining Configuration}\label{sec:pretraining_conifg} 

The training configuration follows the same as Time-MoE: forecasting horizons are set to \(\{1, 8, 32, 64\}\) in the output projection, and the auxiliary loss weighting factor \(\alpha\) is 0.02. We optimize with AdamW using initial learning rate \(1 \times 10^{-3}\), weight decay 0.1, \(\beta_1 = 0.9\), and \(\beta_2 = 0.95\). The learning rate scheduler employs a linear warmup for the first 10,000 steps, followed by cosine annealing to a minimum learning rate of \(5 \times 10^{-5}\). Training is performed on 2 NVIDIA A100 80GB GPUs using BF16 precision, and the configurations for each model are described in detail in Table~\ref{tab:config}.

\begin{table}[ht]
\centering
\caption{Model configurations.}
\label{tab:config}
\scriptsize

\begin{subtable}{\linewidth}
\centering
\begin{tabular}{lccccccc}
\toprule
 & Enc. Layers & Dec. Layers & $d_{\text{model}}$ & $d_{\text{state}}$ & $d_{\text{conv}}$ & expand & Params \\
\midrule
\model$_{\text{base}}$  & 2 & 2 & 384 & 128 & 4 & 4 & 4M \\
\model$_{\text{large}}$ & 2 & 2 & 768 & 128 & 4 & 4 & 16M \\
\bottomrule
\end{tabular}
\caption{Mamba encoder--decoder.}
\label{tab:mamba}
\end{subtable}

\vspace{0.5em} 

\begin{subtable}{\linewidth}
\centering
\begin{tabular}{lcccccccccc}
\toprule
 Model & Layers & Heads & Experts & $K$ & $d_{\text{model}}$ & $d_{\text{ff}}$ & $d_{\text{expert}}$ & Activated Params & Total Params \\
\midrule
\model$_{\text{base}}$  & 12 & 12 & 8 & 2 & 384  & 1536 & 192 & 50M  & 113M \\
\model$_{\text{large}}$ & 12 & 12 & 8 & 2 & 768  & 3072 & 384 & 200M & 453M \\
\bottomrule
\end{tabular}
\caption{Transformer backbone.}
\label{tab:transformer}
\end{subtable}

\end{table}

\section{Downsampling of Pretraining dataset}\label{sec:down}

The original Time-300B dataset is heavily skewed by the Nature domain, which contributed to more than $90\%$ of the dataset, as shown in Table~\ref{tab:time_300b_full}. 

\begin{table}[h]
\centering
\caption{Key statistics of the pre-training dataset Time-300B from various domains.}
\label{tab:time_300b_full}
\scriptsize
\begin{tabular}{lrrrrrrrrrc}
\toprule
& \textbf{Energy} & \textbf{Finance} & \textbf{Healthcare} & \textbf{Nature} & \textbf{Sales} & \textbf{Synthetic} & \textbf{Transport} & \textbf{Web} & \textbf{Other} & \textbf{Total} \\
\midrule
\# Seqs. & 2,875,335 & 1,715 & 1,752 & 31,621,183 & 110,210 & 11,968,625 & 622,414 & 972,158 & 40,265 & 48,220,929 \\
\# Obs. & 15.981 B & 413.696 K & 471.040 K & 279.724 B & 26.382 M & 9.222 B & 2.130 B & 1.804 B & 20.32 M & 309.09 B \\
Percent \% & 5.17\% & 0.0001\% & 0.0001\% & 90.50\% & 0.008\% & 2.98\% & 0.69\% & 0.58\% & 0.006\% & 100\% \\
\bottomrule
\end{tabular}
\end{table}

And within the Nature domain, the 3 largest domains datasets contribute the most, as seen in Table~\ref{tab:nature}.

\begin{table}[h]
\centering
\caption{Key properties of Nature dataset from Time-300B..}
\label{tab:nature}
\footnotesize
\begin{tabular}{|l|c|c|r|r|l|}
\hline
\textbf{Dataset} & \textbf{Domain} & \textbf{Freq.} & \textbf{\# Time Series} & \textbf{\# Obs.} & \textbf{Source} \\
\hline

\rowcolor{orange!30} Weatherbench (Hourly) & Nature & H & 3,984,029 & 74,630,250,518 & \citep{rasp2020weatherbench} \\
Weatherbench (Daily) & Nature & D & 301,229 & 3,223,513,345 & \citep{rasp2020weatherbench} \\
Weatherbench (Weekly) & Nature & W & 226,533 & 462,956,049 & \citep{rasp2020weatherbench} \\
Beijing Air Quality & Nature & H & 4,262 & 2,932,657 & \citep{chen2019beijing} \\
China Air Quality & Nature & H & 17,686 & 4,217,605 & \cite{zheng2015forecasting} \\
\rowcolor{orange!30} CMIP6 & Nature & 6H & 14,327,808 & 104,592,998,400 & \cite{nguyen2023climatelearn} \\
\rowcolor{orange!30} ERA5 & Nature & H & 11,940,789 & 93,768,721,472 & \cite{nguyen2023climatelearn} \\
Oikolab Weather & Nature & H & 309 & 615,574 & \citep{godahewa2021monash} \\
Saugeen & Nature & D & 38 & 17,311 & \citep{godahewa2021monash} \\
Subseasonal & Nature & D & 17,604 & 51,968,498 & \citep{mouatadid2023subseasonalclimateusa} \\
Subseasonal Precipitation & Nature & D & 13,467 & 4,830,284 & \citep{mouatadid2023subseasonalclimateusa} \\
Sunspot & Nature & D & 19 & 45,312 & \citep{godahewa2021monash} \\
Temperature Rain & Nature & D & 13,226 & 3,368,098 & \citep{godahewa2021monash} \\
Weather & Nature & D & 9,525 & 26,036,234 & \citep{ansari2024chronos} \\
\hline
\end{tabular}
\end{table}

In order to reduce the bias from these 3 datasets, we downsample the top 3 datasets by $30 \%$ at random during pretraining, bringing down the total number of samples in the pretraining dataset from 309B to $\approx$120B.

\section{Training Tokens vs Performance}\label{sec:scale}

\begin{figure}[hbt!]
\centering
\includegraphics[width=0.75\textwidth]{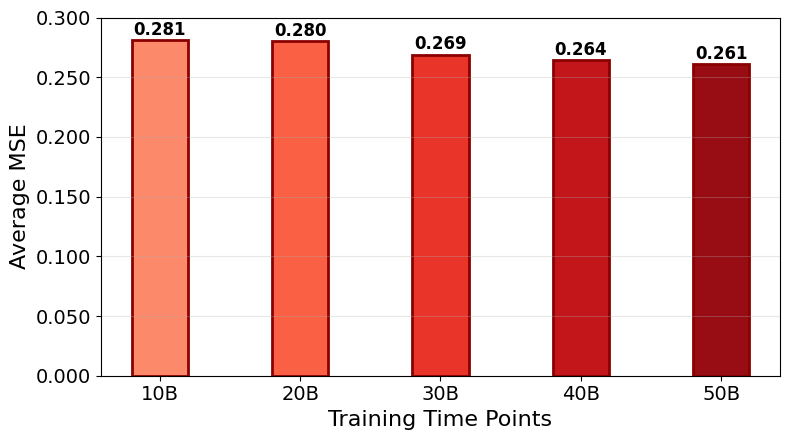}
\caption{Performance scaling with training data size: Average MSE for 96-horizon forecasting across five benchmarks shows consistent improvement with increased training tokens.}
\label{fig:tokens_vs_mse}
\end{figure}

Figure~\ref{fig:tokens_vs_mse} demonstrates that \model exhibits favorable scaling behavior, with performance consistently improving as the training budget increases from 10B to 50B tokens. This scaling trend aligns with observations in~\citep{shi2024time}, indicating that \model can effectively leverage larger datasets and computational resources. The consistent performance gains across different training scales suggest that \model exhibits similar scaling behavior to Time-MoE but with significantly improved data and compute efficiency, positioning it as a promising candidate for even larger-scale pretraining regimes.

\section{Compression Rate vs Performance}\label{sec:rate_vs_mse}

 For the main results, we choose a moderate compression rate of $4 \times$. We now compare the performance against two more variants of \model$_{\text{base}}$ trained with a target compression rate of $6 \times$ and $8 \times$, by adjusting the threshold factor to $0.4$ and $0.45$ respectively. And we plot the average MSE across the five datasets for prediction horizon 96. As expected, while the computational efficiency increases with higher compression, the performance also drops noticeably. techniques such as hierarchical compression~\cite{hwang2025dynamic} might be necessary to alleviate this drop in performance.

 \begin{figure}[hbt!]
\centering
\includegraphics[width=0.75\textwidth]{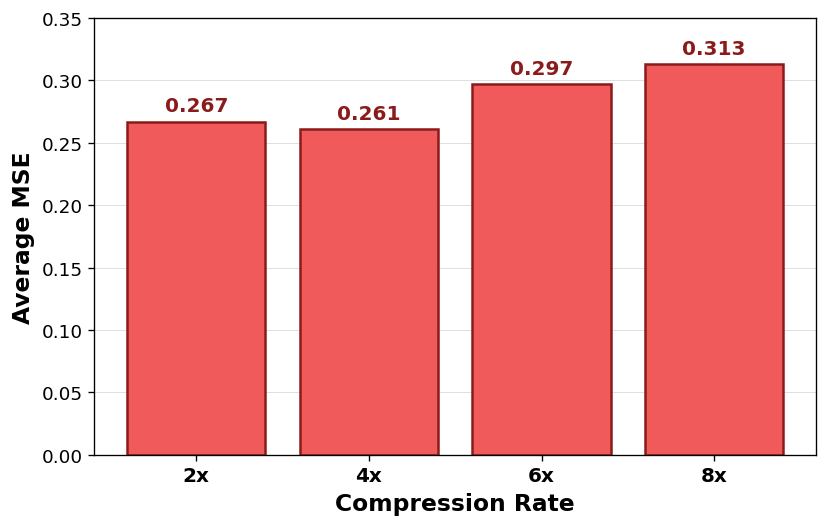}
\caption{Performance scaling with training data size: Average MSE for 96-horizon forecasting across five benchmarks shows consistent improvement with increased training tokens.}
\label{fig:compression_vs_mse}
\end{figure}

\section{Performance for a Fixed Context Length}\label{sec:fixed_context}
\model offers two key advantages over Time-MoE: First is the reduced token count to the Transformer backbone through dynamic compression. Further, \model also improves forecasting capability over longer horizons using shorter historical contexts, compared to point embedding models. 

\begin{figure}[hbt!]
\centering
\includegraphics[width=0.8\textwidth]{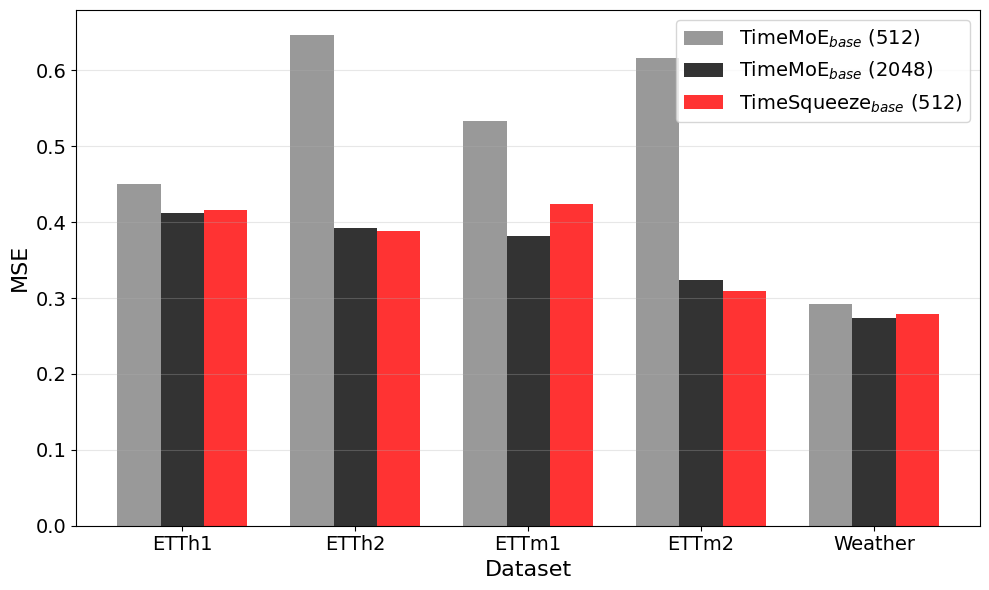}
\caption{Performance comparison between \model and Time-MoE for a given context length for prediction horizon $336$. \model noticeably outperforms the point-embedding baseline when the available context is limited.}
\label{fig:fixed_context}
\end{figure}

Our analysis demonstrates that for a fixed context length, \model significantly outperforms the point embedding baseline Time-MoE when predicting long-horizon forecasts. Figure~\ref{fig:fixed_context} shows that for a given context length of $512$, \model achieves a superior forecasting accuracy for the a horizon of $336$. This improvement stems from our adaptive patching mechanism, which enables the model to extract more informative temporal patterns from limited historical data.


\section{Additional Ablation Results}\label{app:add_ablation}

Table~\ref{tab:main_ablation} contains the full et of results for the ablation studies presented in Section~\ref{sec:ablation}.

\begin{table}[ht]
\raggedright 
\tiny
\caption{Ablation study on zero-shot forecasting performance for prediction horizon 96.}
\setlength{\tabcolsep}{3pt}
\sisetup{table-format=1.4}

\label{tab:main_ablation}
\begin{tabular}{@{}l *{5}{S S} | l l@{}}
\toprule
\multirow{2}{*}{\textbf{Model / Variation}} &
\multicolumn{2}{c}{\textbf{ETTh1}} &
\multicolumn{2}{c}{\textbf{ETTh2}} &
\multicolumn{2}{c}{\textbf{ETTm1}} &
\multicolumn{2}{c}{\textbf{ETTm2}} &
\multicolumn{2}{c}{\textbf{Weather}} &
\multicolumn{2}{c}{\textbf{Average}} \\
\cmidrule(lr){2-3} \cmidrule(lr){4-5} \cmidrule(lr){6-7} \cmidrule(lr){8-9} \cmidrule(lr){10-11} \cmidrule(lr){12-13}
& {\textbf{MSE}} & {\textbf{MAE}} & {\textbf{MSE}} & {\textbf{MAE}} & {\textbf{MSE}} & {\textbf{MAE}} & {\textbf{MSE}} & {\textbf{MAE}} & {\textbf{MSE}} & {\textbf{MAE}} & {\textbf{MSE}} & {\textbf{MAE}} \\
\midrule

\model$_{\text{base}}$ & 0.357 & 0.384 & 0.281 & 0.336 & 0.311 & 0.343 & 0.181 & 0.270 & 0.166 & 0.216 & 0.259 & 0.310 \\
Time-MoE$_{\text{base}}$        & 0.357 & 0.381 & 0.305 & 0.359 & 0.338 & 0.368 & 0.201 & 0.291 & 0.160 & 0.214 & 0.272 (+5.0\%) & 0.323 (+4.2\%) \\
\midrule
\addlinespace

\model w/ fixed patching (size 4)     & 0.373 & 0.396 & 0.455 & 0.448 & 0.359 & 0.382 & 0.335 & 0.380 & 0.178 & 0.232 & 0.340 (+31.3\%) & 0.368 (+18.7\%) \\

\model w/ fixed patching (size 2) & 0.376 & 0.402 & 0.475 & 0.462 & 0.361 & 0.384 & 0.357 & 0.396 & 0.179 & 0.233 & 0.340 (+31.3\%) & 0.368 (+18.7\%) \\

\model w/ linear patching (no SSM)       & 0.379 & 0.401 & 0.481 & 0.463 & 0.370 & 0.375 & 0.375 & 0.402 & 0.158 & 0.174 & 0.353 (+36.3\%) & 0.363 (+17.1\%) \\

\model w/o fine-grained features      & 0.366 & 0.388 & 0.277 & 0.342 & 0.339 & 0.362 & 0.187 & 0.283 & 0.169 & 0.218 & 0.268 (+3.5\%) & 0.319 (+2.9\%) \\

\model w/o original pos. IDs  & 0.375 & 0.393 & 0.291 & 0.358 & 0.346 & 0.363 & 0.191 & 0.293 & 0.169 & 0.219 & 0.274 (+5.8\%) & 0.325 (+4.8\%) \\

\bottomrule
\end{tabular}
\end{table}

\subsection{Inference Context Length vs Performance}\label{sec:inference_short}

While longer context lengths generally provide more historical information for forecasting, the relationship between context length and performance is not monotonic. We investigate the effect of varying inference context lengths on forecasting accuracy by evaluating \model with context lengths ranging from 96 to 1536
tokens while keeping all other hyperparameters fixed.

\begin{figure}[hbt!]
\centering
\includegraphics[width=0.7\textwidth]{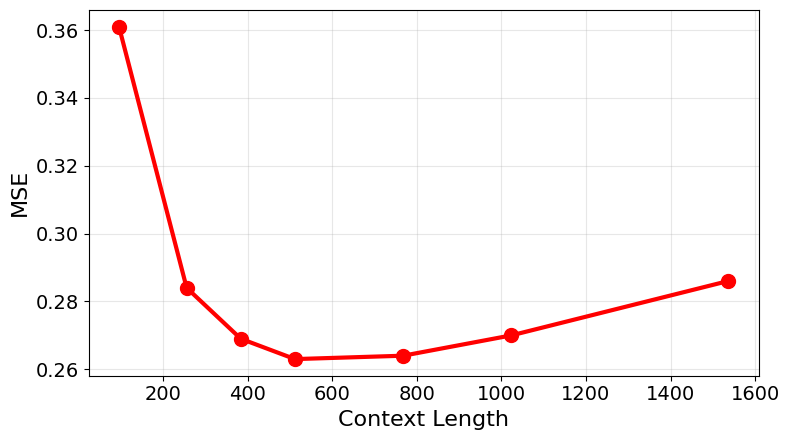}
\caption{Forecasting performance improves with context length up to 512-800 tokens, then plateaus or slightly degrades.}
\label{fig:inf_length_vs_performance}
\end{figure}

Figure~\ref{fig:inf_length_vs_performance} reveals that performance initially improves as context length increases from 96 to 1536, reaching optimal performance around 512. However, further increasing the context length beyond this range leads to marginal performance degradation. This suggests that while additional historical context can be beneficial up to a certain point, excessively long contexts may introduce noise or make it harder for the model to focus on the most relevant patterns. 

\subsection{Generalization of gains: non-MoE decoder only backbone}\label{sec:gen_backbone}

Given that Time-MoE uses a generic decoder-only backbone + mixture of experts, and that dynamic patching operates purely at the input representation level, we expect the observed improvements to transfer to other modern architectures as well. Exploring the pretraining of these additional backbones at scale is a valuable direction for future work.

As preliminary evidence of generality, we replaced the entire Time-MoE-small backbone with a generic 10M-parameter decoder-only Transformer and conducted a controlled ablation between fixed patching and dynamic patching. We stick to the same pretraining context length of 2048 and inference context length of 512 and test the performance for a prediction horizon of 96. This experiment removes any reliance on architectural details unique to Time-MoE and evaluates the patching mechanism on a completely different, lightweight backbone. As shown in the table below, dynamic patching still outperforms fixed patching even under this generic architecture, further supporting that the benefits of dynamic within-series patching arise from the method itself rather than from Time-MoE-specific design choices.

\begin{table}[H]
\centering
\resizebox{\textwidth}{!}{%
\begin{tabular}{lccccc}
\toprule
\textbf{Method} & \textbf{ETTh1 (MSE/MAE)} & \textbf{ETTh2 (MSE/MAE)} & \textbf{ETTm1 (MSE/MAE)} & \textbf{ETTm2 (MSE/MAE)} & \textbf{Weather (MSE/MAE)} \\
\midrule
Dynamic patching (avg patch size 4) & 0.342 / 0.364 & 0.280 / 0.347 & 0.351 / 0.370 & 0.201 / 0.292 & 0.175 / 0.229 \\
Fixed-size patching (size 4)        & 0.366 / 0.394 & 0.406 / 0.421 & 0.370 / 0.388 & 0.362 / 0.406 & 0.185 / 0.250 \\
\bottomrule
\end{tabular}%
}
\caption{Evaluation of TimeSqueeze Dynamic patching with a generic transformer backbone.}
\label{tab:patching_results}
\end{table}

\subsection{Generalization of gains: Pretraining on GiftEvalPretrain dataset}\label{sec:gen_dataset}

We now use the same 10M parameter variant of TimeSqueeze to study the effect of pretraining dataset on zero-shot forecasting. We consider 
GiftEvalPretrain~\cite{aksu2024giftevalbenchmarkgeneraltime}, a popular pretraining dataset for time series forecasting. Table~\ref{tab:forecast_gifteval} shows that \model with dynamic patching consistently outperforms equivalent fixed patching and point embedding baselines.

\begin{table*}[t]
\centering
\scriptsize
\renewcommand{\arraystretch}{1.0}
\setlength{\tabcolsep}{4pt}
\caption{Performance of TimeSqueeze-10M variant with pretraining of GiftEvalPretrain dataset, showing consistent gains of dynamic patching over fixed patching.}
\label{tab:forecast_gifteval}
\resizebox{0.95\textwidth}{!}{
\begin{tabular}{@{}l *{5}{S S} | S S@{}}
\toprule
\multicolumn{1}{c}{\multirow{2}{*}{\textbf{Method}}} & \multicolumn{2}{c}{\textbf{ETTh1}} & \multicolumn{2}{c}{\textbf{ETTh2}} & \multicolumn{2}{c}{\textbf{ETTm1}} & \multicolumn{2}{c}{\textbf{ETTm2}} & \multicolumn{2}{c|}{\textbf{Weather}} & \multicolumn{2}{c}{\textbf{Avg.}} \\
\cmidrule(lr){2-3} \cmidrule(lr){4-5} \cmidrule(lr){6-7} \cmidrule(lr){8-9} \cmidrule(lr){10-11} \cmidrule(lr){12-13}
& \textbf{MSE} & \textbf{MAE} & \textbf{MSE} & \textbf{MAE} & \textbf{MSE} & \textbf{MAE} & \textbf{MSE} & \textbf{MAE} & \textbf{MSE} & \textbf{MAE} & \textbf{MSE} & \textbf{MAE} \\
\midrule
Dynamic patching & \textbf{0.369} & \textbf{0.396} & \textbf{0.364} & \textbf{0.400} & \textbf{0.411} & \textbf{0.415} & \textbf{0.287} & \textbf{0.363} & \textbf{0.204} & \textbf{0.265} & \textbf{0.327} & \textbf{0.368} \\
Fixed-size patching & 0.382 & 0.403 & 0.405 & 0.428 & 0.439 & 0.424 & 0.334 & 0.394 & 0.209 & 0.265 & 0.354 & 0.383 \\
Point embedding & 0.392 & 0.415 & 0.567 & 0.506 & 0.426 & 0.441 & 0.660 & 0.548 & 0.235 & 0.302 & 0.456 & 0.442 \\
\bottomrule
\end{tabular}
}
\end{table*}

\section{Additional Forecasting Results}\label{app:add_ablation}

\subsection{Complete results for Table ~\ref{tab:results_full_shot_time_moe_avg}}\label{sec:full_shot_full}

The forecasting results at per prediction horizon from ~\secref{sec:full_shot_avg} are provided in Table~\ref{tab:results_full_shot_time_moe}.

\begin{table*}
\centering
\scriptsize 
\renewcommand{\arraystretch}{1.0} 
\setlength{\tabcolsep}{2pt} 
\sisetup{table-format=1.3} 
\caption{Performance comparison of full-shot forecasting. \textbf{Bold} for best and \underline{underscore} for 2nd best. 
}
\vspace{1em}
\label{tab:results_full_shot_time_moe}
\resizebox{0.95\textwidth}{!}{
\begin{tabular}{@{}ll *{9}{S S}@{}}
\toprule
\multicolumn{1}{c}{\multirow{2}{*}{\textbf{Models}}} & \multicolumn{1}{c|}{\multirow{2}{*}{\textbf{Metrics}}} & \multicolumn{2}{c}{\textbf{\model$_{\text{base}}$}} & \multicolumn{2}{c}{\textbf{Time-MoE$_{\text{base}}$}} & \multicolumn{2}{c}{\textbf{iTransformer}} & \multicolumn{2}{c}{\textbf{TimeMixer}} & \multicolumn{2}{c}{\textbf{TimesNet}} & \multicolumn{2}{c}{\textbf{PatchTST}} & \multicolumn{2}{c}{\textbf{DLinear}} & \multicolumn{2}{c}{\textbf{CycleNet$_{\text{MLP}}$}} & \multicolumn{2}{c}{\textbf{TQNet}} \\
\cmidrule(lr){3-4} \cmidrule(lr){5-6} \cmidrule(lr){7-8} \cmidrule(lr){9-10} \cmidrule(lr){11-12} \cmidrule(lr){13-14} \cmidrule(lr){15-16} \cmidrule(lr){17-18} \cmidrule(lr){19-20}
& & \textbf{MSE} & \textbf{MAE} & \textbf{MSE} & \textbf{MAE} & \textbf{MSE} & \textbf{MAE} & \textbf{MSE} & \textbf{MAE} & \textbf{MSE} & \textbf{MAE} & \textbf{MSE} & \textbf{MAE} & \textbf{MSE} & \textbf{MAE} & \textbf{MSE} & \textbf{MAE} & \textbf{MSE} & \textbf{MAE} \\
\midrule
\multirow{5}{*}{ETTh1} & 96 & \underline{0.354} & \underline{0.384} & \textbf{0.345} & \textbf{0.375} & 0.386 & 0.405 & 0.375 & 0.400 & 0.384 & 0.402 & 0.414 & 0.419 & 0.423 & 0.448 & 0.375 & 0.395 & 0.371 & 0.393 \\
 & 192 & \underline{0.397} & \underline{0.412} & \textbf{0.372} & \textbf{0.396} & 0.441 & 0.436 & 0.436 & 0.429 & 0.421 & 0.429 & 0.460 & 0.445 & 0.471 & 0.474 & 0.436 & 0.428 & 0.428 & 0.426 \\
 & 336 & \underline{0.418} & \underline{0.427} & \textbf{0.389} & \textbf{0.412} & 0.487 & 0.458 & 0.484 & 0.458 & 0.491 & 0.469 & 0.501 & 0.466 & 0.570 & 0.546 & 0.496 & 0.455 & 0.476 & 0.446 \\
 & 720 & \underline{0.423} & \underline{0.454} & \textbf{0.410} & \textbf{0.443} & 0.503 & 0.491 & 0.498 & 0.482 & 0.521 & 0.500 & 0.500 & 0.488 & 0.653 & 0.621 & 0.520 & 0.484 & 0.487 & 0.470 \\
\rowcolor{gray!20} & \textbf{Avg.} & \underline{0.398} & \underline{0.419} & \textbf{0.379} & \textbf{0.406} & 0.454 & 0.447 & 0.448 & 0.442 & 0.454 & 0.450 & 0.468 & 0.454 & 0.529 & 0.522 & 0.457 & 0.441 & 0.441 & 0.434 \\
\hline \\[-0.5em]
\multirow{5}{*}{ETTh2} & 96 & \textbf{0.274} & \textbf{0.336} & \underline{0.276} & \underline{0.340} & 0.297 & 0.349 & 0.289 & 0.341 & 0.340 & 0.374 & 0.302 & 0.348 & 0.745 & 0.584 & 0.298 & 0.344 & 0.295 & 0.343 \\
 & 192 & \underline{0.337} & \underline{0.379} & \textbf{0.331} & \textbf{0.371} & 0.380 & 0.400 & 0.372 & 0.392 & 0.402 & 0.414 & 0.388 & 0.400 & 0.877 & 0.656 & 0.372 & 0.396 & 0.367 & 0.393 \\
 & 336 & \textbf{0.373} & \underline{0.408} & \textbf{0.373} & \textbf{0.402} & 0.428 & 0.432 & \underline{0.386} & 0.414 & 0.452 & 0.541 & 0.426 & 0.433 & 1.043 & 0.731 & 0.431 & 0.439 & 0.417 & 0.427 \\
 & 720 & 0.417 & 0.449 & \textbf{0.404} & \textbf{0.431} & 0.427 & 0.445 & \underline{0.412} & \underline{0.434} & 0.462 & 0.657 & 0.431 & 0.446 & 1.104 & 0.763 & 0.450 & 0.458 & 0.433 & 0.446 \\
\rowcolor{gray!20} & \textbf{Avg.} & \underline{0.350} & \underline{0.393} & \textbf{0.346} & \textbf{0.386} & 0.383 & 0.406 & 0.364 & 0.395 & 0.414 & 0.496 & 0.386 & 0.406 & 0.942 & 0.683 & 0.388 & 0.409 & 0.378 & 0.402 \\
\hline \\[-0.5em]
\multirow{5}{*}{ETTm1} & 96 & \underline{0.289} & \textbf{0.332} & \textbf{0.286} & \underline{0.334} & 0.334 & 0.368 & 0.320 & 0.357 & 0.338 & 0.375 & 0.329 & 0.367 & 0.404 & 0.426 & 0.319 & 0.360 & 0.311 & 0.353 \\
 & 192 & \underline{0.344} & \underline{0.366} & \textbf{0.307} & \textbf{0.358} & 0.377 & 0.391 & 0.360 & 0.381 & 0.374 & 0.387 & 0.367 & 0.385 & 0.450 & 0.451 & 0.360 & 0.381 & 0.356 & 0.378 \\
 & 336 & 0.398 & \underline{0.396} & \textbf{0.354} & \textbf{0.390} & 0.426 & 0.420 & 0.390 & 0.404 & 0.410 & 0.411 & 0.399 & 0.410 & 0.532 & 0.515 & \underline{0.389} & 0.403 & 0.390 & 0.401 \\
 & 720 & 0.502 & 0.451 & \textbf{0.433} & 0.445 & 0.491 & 0.459 & 0.454 & 0.441 & 0.478 & 0.450 & 0.454 & \textbf{0.439} & 0.666 & 0.589 & \underline{0.447} & 0.441 & 0.452 & \underline{0.440} \\
\rowcolor{gray!20} & \textbf{Avg.} & 0.383 & \underline{0.386} & \textbf{0.345} & \textbf{0.381} & 0.407 & 0.409 & 0.381 & 0.395 & 0.400 & 0.405 & 0.387 & 0.400 & 0.513 & 0.495 & 0.379 & 0.396 & \underline{0.377} & 0.393 \\
\hline \\[-0.5em]
\multirow{5}{*}{ETTm2} & 96 & \underline{0.168} & \underline{0.256} & 0.172 & 0.265 & 0.180 & 0.264 & 0.175 & 0.258 & 0.187 & 0.267 & 0.175 & 0.259 & 0.287 & 0.366 & \textbf{0.163} & \textbf{0.246} & 0.173 & \underline{0.256} \\
 & 192 & \textbf{0.225} & \underline{0.298} & \underline{0.228} & 0.306 & 0.250 & 0.309 & 0.237 & 0.299 & 0.249 & 0.309 & 0.241 & 0.302 & 0.414 & 0.392 & 0.229 & \textbf{0.290} & 0.238 & \underline{0.298} \\
 & 336 & \textbf{0.278} & \underline{0.335} & \underline{0.281} & 0.345 & 0.311 & 0.348 & 0.298 & 0.340 & 0.321 & 0.351 & 0.305 & 0.343 & 0.597 & 0.542 & 0.284 & \textbf{0.327} & 0.301 & 0.340 \\
 & 720 & \textbf{0.366} & \underline{0.395} & 0.403 & 0.424 & 0.412 & 0.407 & 0.391 & 0.396 & 0.408 & 0.403 & 0.402 & 0.400 & 1.730 & 1.042 & \underline{0.389} & \textbf{0.391} & 0.397 & 0.396 \\
\rowcolor{gray!20} & \textbf{Avg.} & \textbf{0.259} & \underline{0.321} & 0.271 & 0.335 & 0.288 & 0.332 & 0.275 & 0.323 & 0.291 & 0.332 & 0.280 & 0.326 & 0.757 & 0.610 & \underline{0.266} & \textbf{0.314} & 0.277 & 0.323 \\
\hline \\[-0.5em]  
\multirow{5}{*}{Weather} & 96 & \underline{0.152} & \textbf{0.199} & \textbf{0.151} & 0.203 & 0.154 & 0.208 & 0.163 & 0.209 & 0.172 & 0.220 & 0.177 & 0.218 & 0.158 & 0.230 & 0.158 & 0.203 & 0.157 & \underline{0.200} \\
 & 192 & \underline{0.201} & 0.249 & \textbf{0.195} & \underline{0.246} & 0.202 & 0.251 & 0.208 & 0.250 & 0.219 & 0.261 & 0.225 & 0.259 & 0.206 & 0.277 & 0.207 & 0.247 & 0.206 & \textbf{0.245} \\
 & 336 & 0.259 & 0.297 & \textbf{0.247} & \underline{0.288} & 0.252 & \textbf{0.287} & \underline{0.251} & \textbf{0.287} & 0.280 & 0.306 & 0.278 & 0.297 & 0.272 & 0.335 & 0.262 & 0.289 & 0.262 & \textbf{0.287} \\
 & 720 & 0.360 & 0.372 & 0.352 & 0.366 & \textbf{0.302} & 0.376 & 0.339 & \textbf{0.341} & 0.365 & 0.359 & 0.354 & 0.348 & \underline{0.308} & 0.418 & 0.344 & 0.344 & 0.344 & \underline{0.342} \\
\rowcolor{gray!20} & \textbf{Avg.} & 0.243 & 0.279 & \textbf{0.236} & 0.275 & 0.250 & 0.280 & \underline{0.240} & \underline{0.271} & 0.259 & 0.286 & 0.258 & 0.280 & 0.258 & 0.315 & 0.243 & \underline{0.271} & 0.242 & \textbf{0.269} \\
\midrule
\rowcolor{gray!20}\textbf{Average} &  & \underline{0.327} & \underline{0.360} & \textbf{0.315} & \textbf{0.357} & 0.356 & 0.375 & 0.342 & 0.365 & 0.364 & 0.394 & 0.356 & 0.373 & 0.600 & 0.525 & 0.347 & 0.366 & 0.343 & 0.364 \\
\midrule
\bottomrule
\end{tabular}
}
\end{table*}

\section{Visualization of patching}\label{sec:vis_patches}

\begin{figure}[htbp]
    \centering
    \begin{subfigure}{\linewidth}
        \centering
        \includegraphics[width=1.0\linewidth]{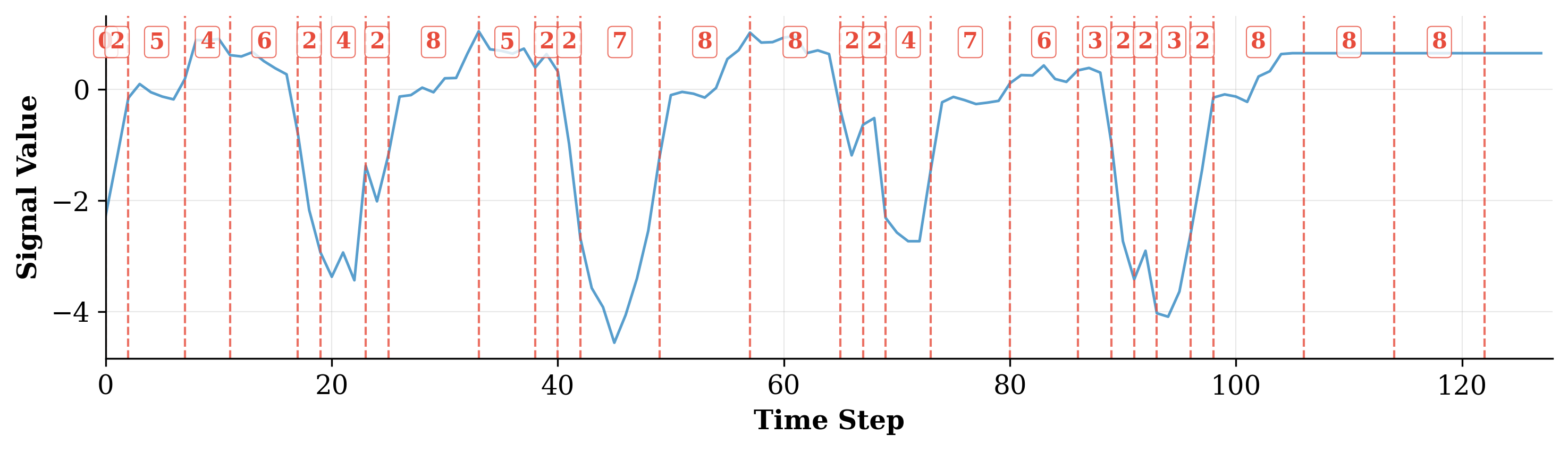}
        \caption{ETTh1 }
        \label{fig:patch-etth1}
    \end{subfigure}
    \vspace{1em}
    \begin{subfigure}{\linewidth}
        \centering
        \includegraphics[width=1.0\linewidth]{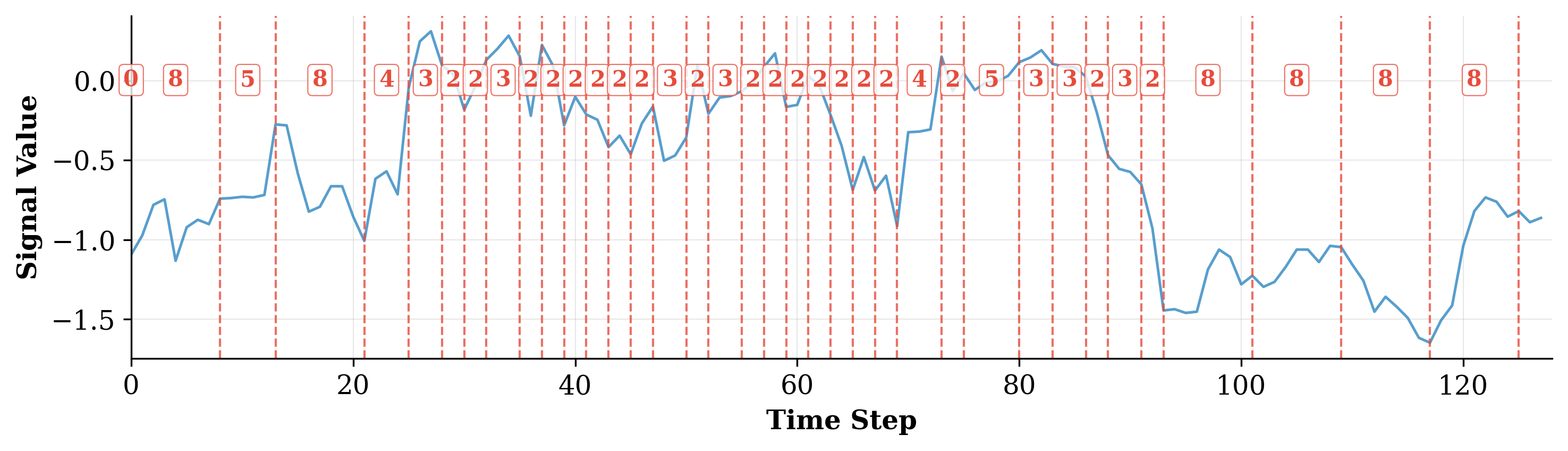}
        \caption{ETTh2 }
        \label{fig:patch-etth2}
    \end{subfigure}
    \caption{Example dynamic patch boundaries for ETTh1 and ETTh2 datasets. }
    \label{fig:etth-fig}
\end{figure}

\begin{figure}[htbp]
    \centering
    \begin{subfigure}{\linewidth}
        \centering
        \includegraphics[width=1.0\linewidth]{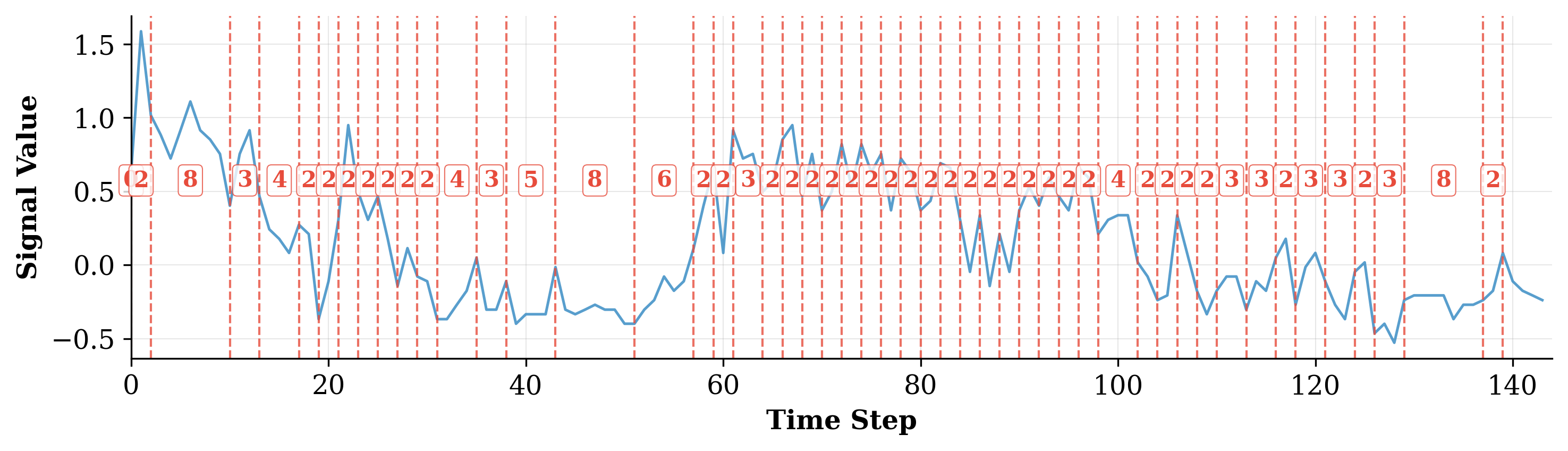}
        \caption{ETTm1 }
        \label{fig:patch-ettm1}
    \end{subfigure}
    \vspace{1em}
    \begin{subfigure}{\linewidth}
        \centering
        \includegraphics[width=1.0\linewidth]{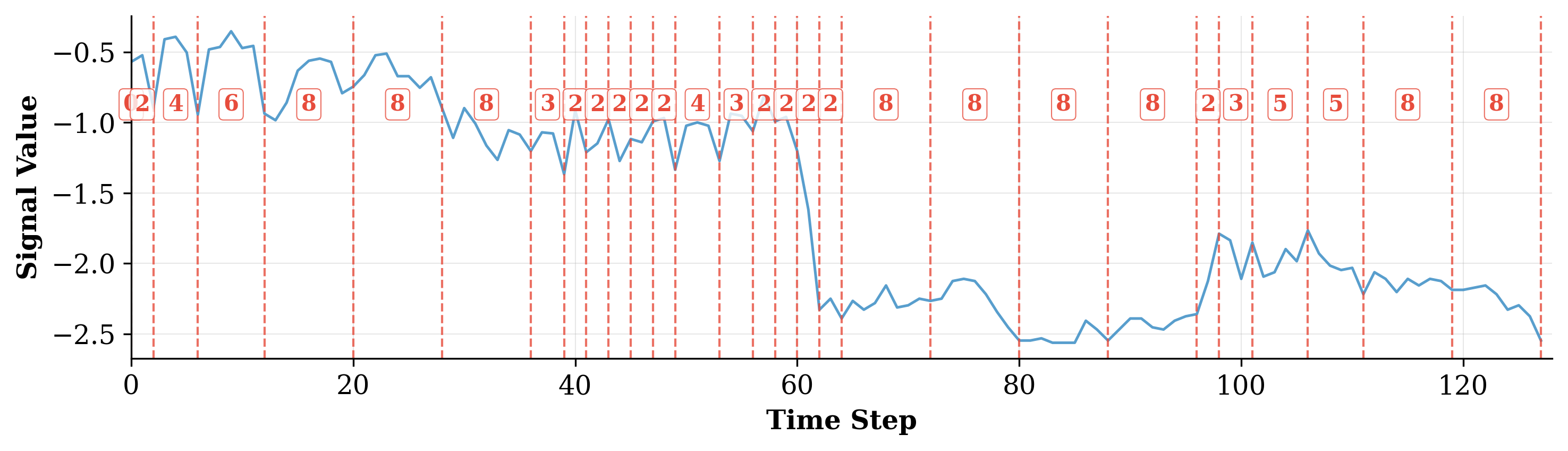}
        \caption{ETTm2 }
        \label{fig:patch-ettm2}
    \end{subfigure}
    \caption{Example dynamic patch boundaries for ETTm1 and ETTm2 datasets. }
    \label{fig:ettm-fig}
\end{figure}

\begin{figure}[htbp]
    \centering
    \begin{subfigure}{\linewidth}
        \centering
        \includegraphics[width=1.0\linewidth]{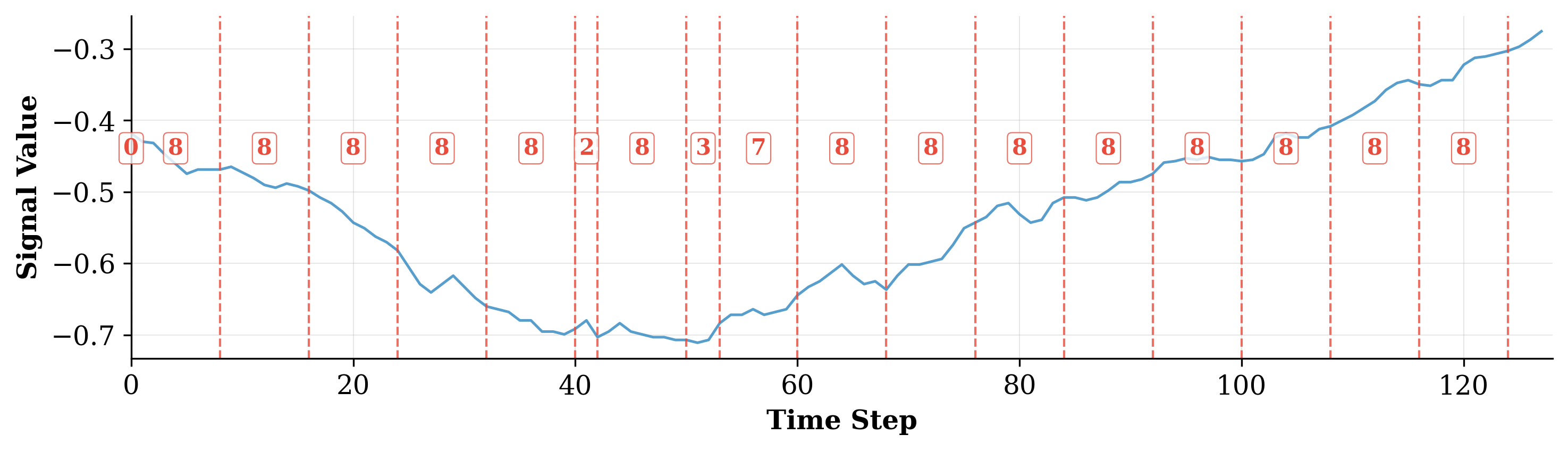}
        \caption{Weather }
        \label{fig:patch-weather}
    \end{subfigure}
    \caption{Example dynamic patch boundaries for Weather dataset. }
    \label{fig:weather-fig}
\end{figure}

We provide the visualization of dynamic patches computed for an example segment of 128 samples from each of the evaluation datasets in Figures~\ref{fig:etth-fig}, ~\ref{fig:ettm-fig}, and~\ref{fig:weather-fig}. As we can see, weather dataset has slower variation in data resulting in larger patch sizes, whereas ETTm data has several regions with rapidly varying signal, resulting in much smaller patch sizes.

\section{Patch distribution}\label{sec:vis_patches}

\begin{figure}[hbt!]
\centering
\includegraphics[width=1.0\textwidth]{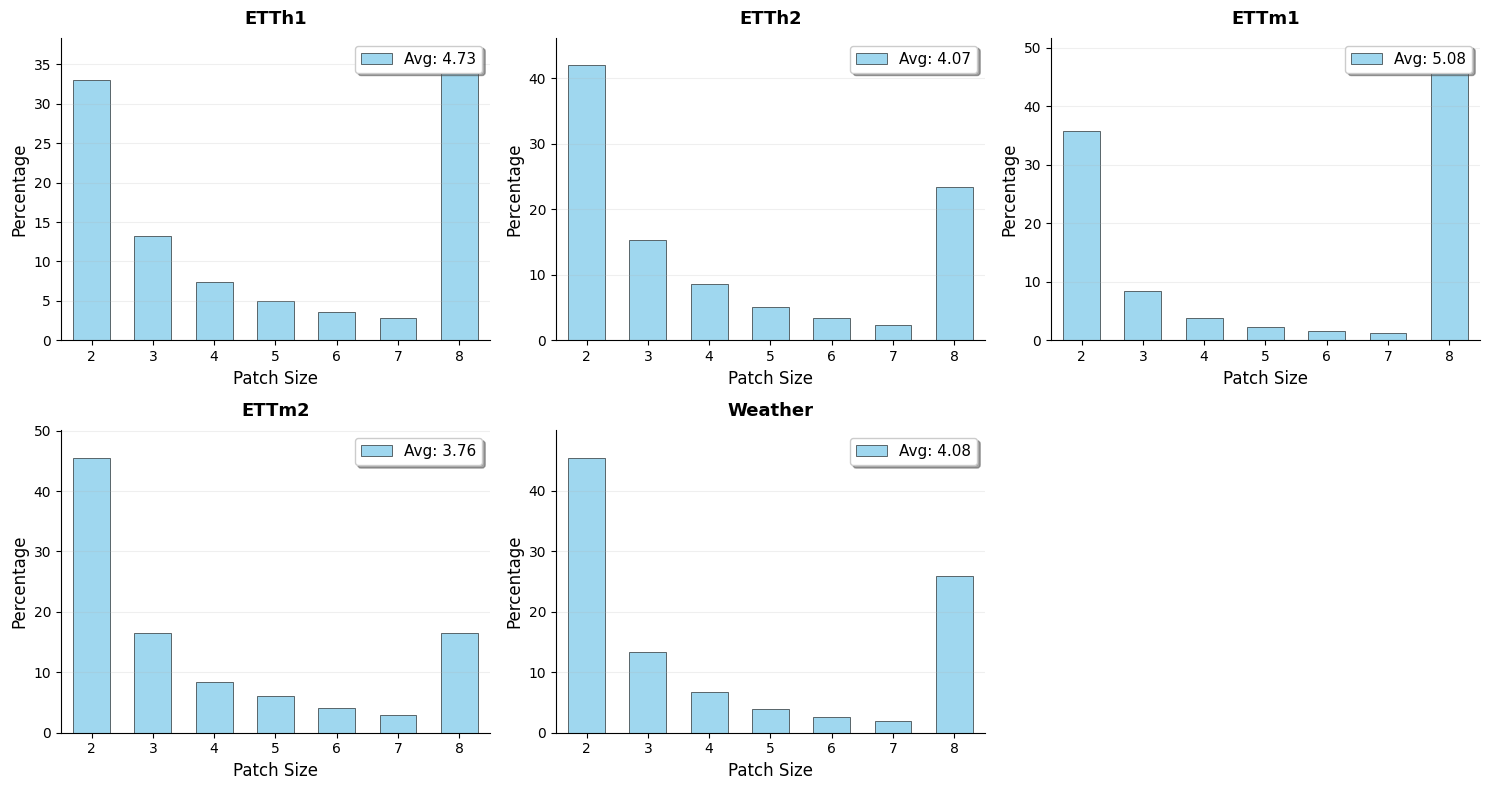}
\caption{Distribution of patch sizes across eval datasets.}
\label{fig:patch_hist}
\end{figure}

We provide the visualization of patch size distributions for each of the eval datasets in Figures~\ref{fig:patch_hist}. As we can see, weather dataset has slower variation in data resulting in larger avg patch size, whereas ETTm2 data has several regions with rapidly varying signal, resulting in much smaller patch sizes.

\section{Visualization of forecasts}\label{sec:vis_forecasts}

\begin{figure}[htbp]
    \centering
    \begin{subfigure}{\linewidth}
        \centering
        \includegraphics[width=1.0\linewidth]{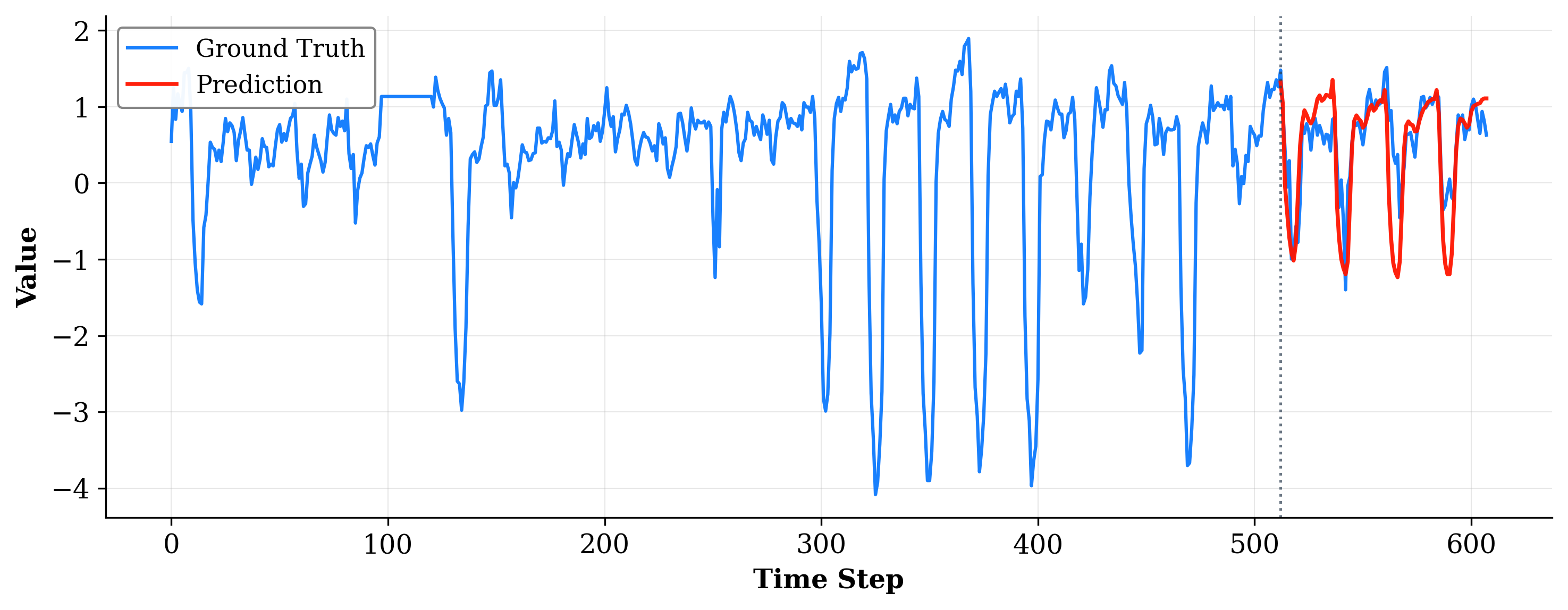}
        \caption{ETTh1}
        \label{fig:forecast-etth1}
    \end{subfigure}
    \vspace{1em}
    \begin{subfigure}{\linewidth}
        \centering
        \includegraphics[width=1.0\linewidth]{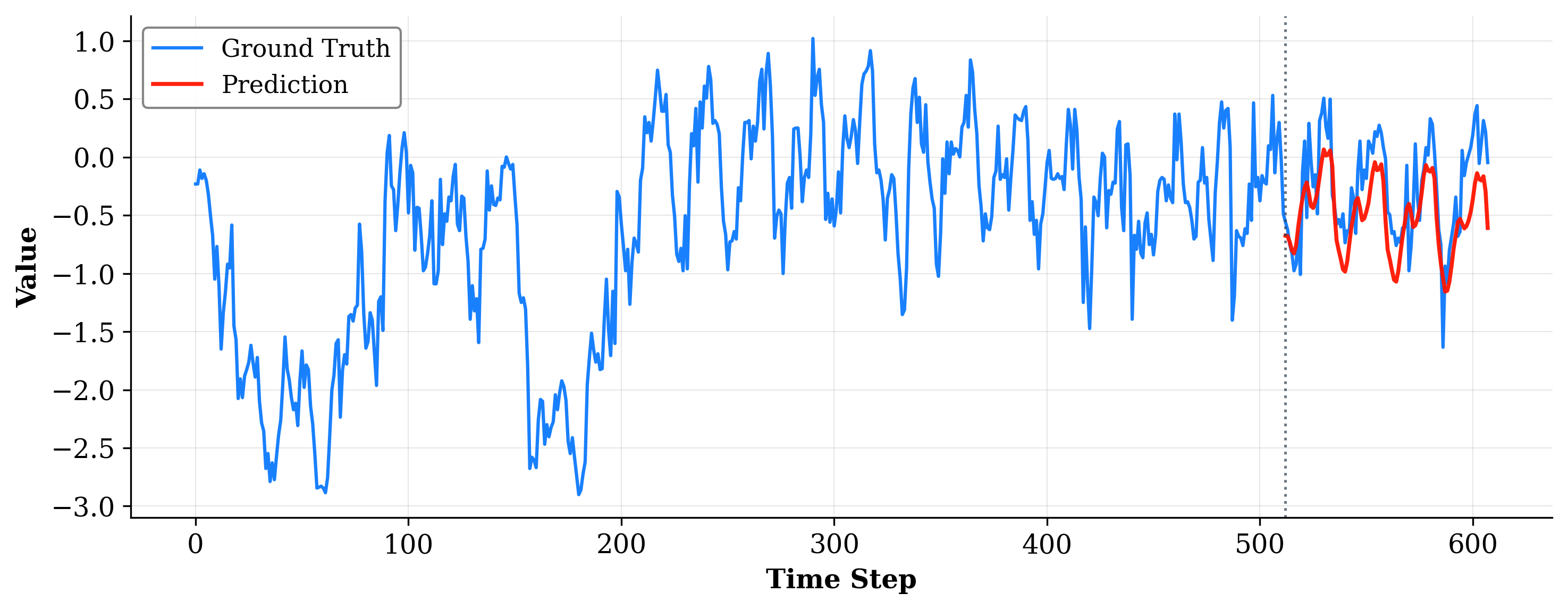}
        \caption{ETTh2}
        \label{fig:forecast-etth2}
    \end{subfigure}
    \caption{Forecasting results on ETTh1 and ETTh2 datasets.}
    \label{fig:forecast-etth}
\end{figure}

\begin{figure}[htbp]
    \centering
    \begin{subfigure}{\linewidth}
        \centering
        \includegraphics[width=1.0\linewidth]{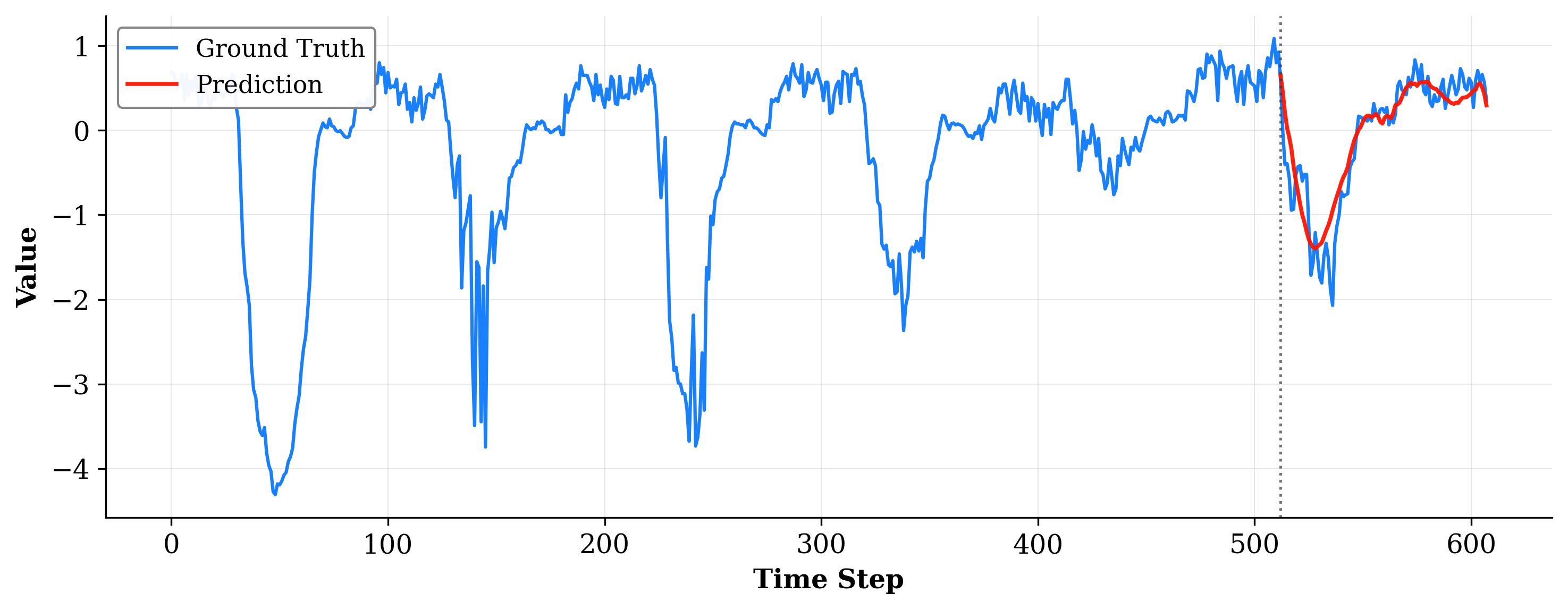}
        \caption{ETTm1}
        \label{fig:forecast-ettm1}
    \end{subfigure}
    \vspace{1em}
    \begin{subfigure}{\linewidth}
        \centering
        \includegraphics[width=1.0\linewidth]{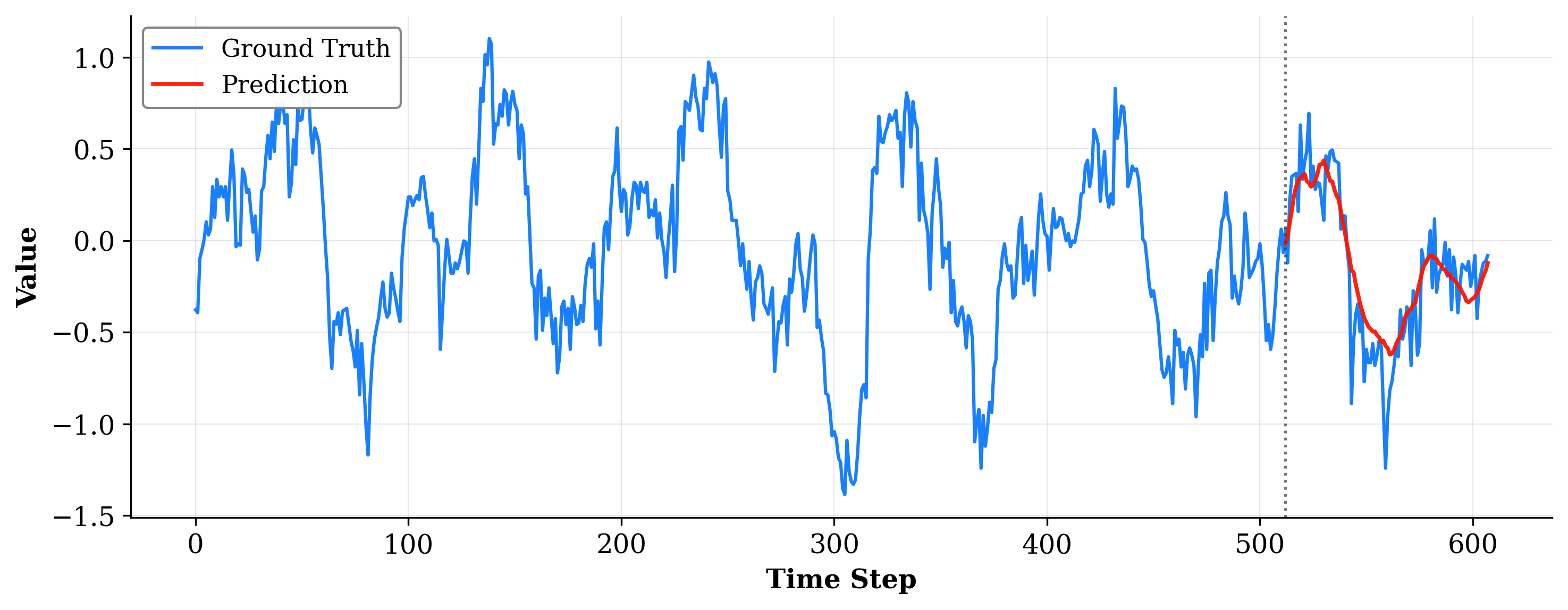}
        \caption{ETTm2}
        \label{fig:forecast-ettm2}
    \end{subfigure}
    \caption{Forecasting results on ETTm1 and ETTm2 datasets.}
    \label{fig:forecast-ettm}
\end{figure}

\begin{figure}[htbp]
    \centering
    \begin{subfigure}{\linewidth}
        \centering
        \includegraphics[width=1.0\linewidth]{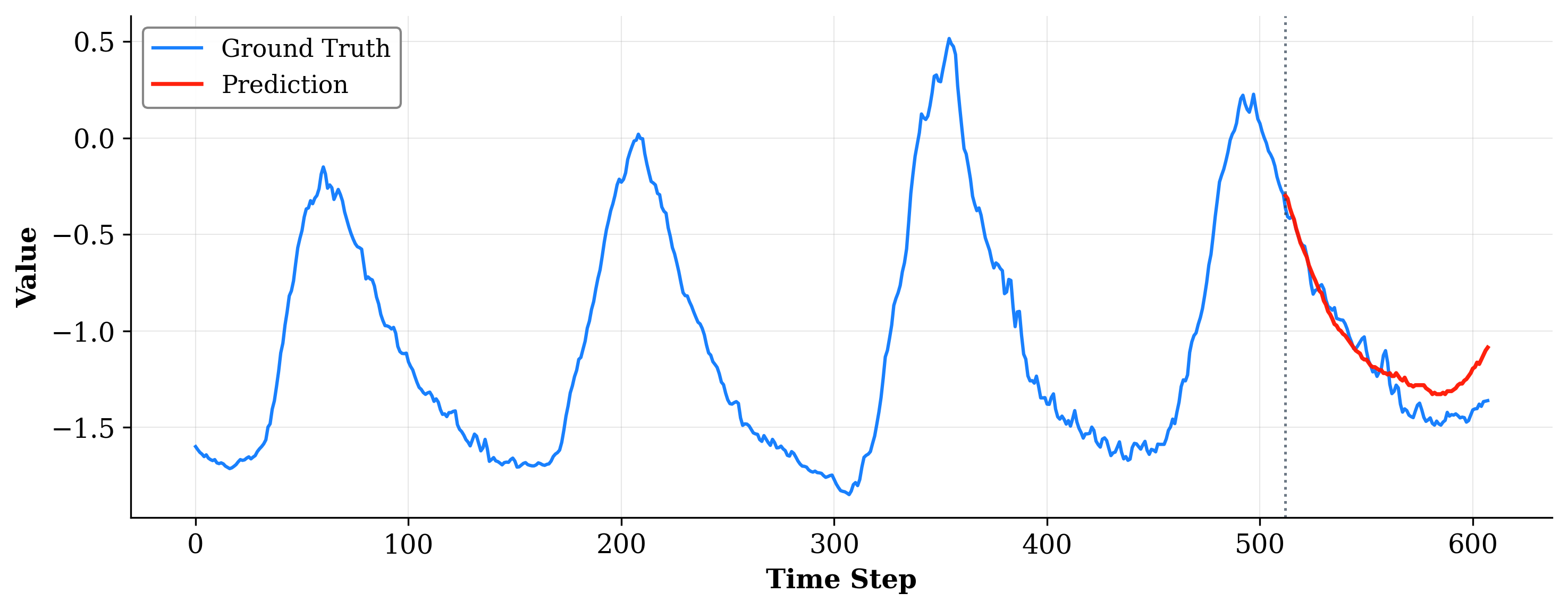}
        \caption{Weather}
        \label{fig:forecast-weather}
    \end{subfigure}
    \caption{Forecasting results on the Weather dataset.}
    \label{fig:forecast-weather-fig}
\end{figure}

We provide the visualization of forecasting results for an example segment of 128 samples from each of the evaluation datasets in Figures~\ref{fig:forecast-etth}, \ref{fig:forecast-ettm}, and~\ref{fig:forecast-weather-fig}. As observed, the Weather dataset exhibits relatively smooth and slowly varying dynamics, making forecasts easier to capture, whereas the ETTm datasets contain regions with rapid fluctuations, which pose greater challenges for accurate prediction.

\end{document}